\documentclass{article}

\usepackage{arxiv}

\usepackage[utf8]{inputenc} %
\usepackage[T1]{fontenc}    %
\usepackage{hyperref}       %
\usepackage{url}            %
\usepackage{booktabs}       %
\usepackage{amsfonts}       %
\usepackage{nicefrac}       %
\usepackage{microtype}      %
\usepackage{lipsum}
\usepackage{graphicx}
\usepackage{makecell}

\usepackage{xspace}
\usepackage{pifont}
\usepackage{hhline}
\usepackage{amsmath, amsthm}
\usepackage[dvipsnames, table]{xcolor}
\usepackage{accents}
\usepackage[numbers,sort]{natbib}
\usepackage{subcaption}
\usepackage{color, colortbl}

\hypersetup{
     colorlinks   = true,
     citecolor    = Blue,
     urlcolor     = Magenta,
     linkcolor    = Blue
}

\definecolor{LightRed}{rgb}{1,0.4,0.4}
\definecolor{LightGreen}{rgb}{0,0.8,0.4}

\newcommand*{\Comb}[2]{{}^{#1}C_{#2}}%

\usepackage[ruled,linesnumbered]{algorithm2e}
\SetKwComment{Comment}{$\triangleright$\ }{}

\hypersetup{
     colorlinks   = true,
     citecolor    = Blue,
     urlcolor     = Magenta,
     linkcolor    = Blue
}

\usepackage{misc_symbols}

\newcommand{\itbf}[1]{ \textit{\textbf{#1}} } %

\newcommand{\robust}{robust\xspace}
\newcommand{\robustness}{robustness\xspace}
\newcommand{\nonrobust}{non-robust\xspace}

\newcommand{\Robustness}{Robustness\xspace}

\newcommand{\concept}{\text{MACC}\xspace}
\newcommand{\concepts}{\text{MACCs}\xspace}

\newcommand{\conceptlabel}{\concept}
\newcommand{\conceptlabels}{\concepts}

\newcommand{\rsn}{\text{con}\xspace}

\title{Unifying Model Explainability and \Robustness\\
via Machine-Checkable Concepts}
\author{
Vedant Nanda \\
 MPI-SWS \& \\ University of Maryland \\
 \texttt{vedant@cs.umd.edu}
 \And
Till Speicher \\
  MPI-SWS\\
  \texttt{tspeicher@mpi-sws.org} \\
  \And
John P. Dickerson \\
  University of Maryland \\
  \texttt{john@cs.umd.edu} \\
  \And
Krishna P. Gummadi \\
  MPI-SWS \\
  \texttt{gummadi@mpi-sws.org} \\
  \And
Muhammad Bilal Zafar \\
  Bosch Center for Artificial Intelligence \\
  \texttt{muhammadbilal.zafar@de.bosch.com} \\
 }

\begin{document}

\maketitle

\begin{abstract}
    As deep neural networks (DNNs) get adopted in an ever-increasing number of applications, explainability has emerged as a crucial desideratum for these models.
    In many real-world tasks, one of the principal reasons for requiring explainability is to in turn assess prediction \robustness, where predictions (\ie, class labels) that do not conform to their respective explanations (\eg, presence or absence of a concept in the input) are deemed to be unreliable.
    However, most, if not all, prior methods for checking explanation-conformity (\eg, LIME, TCAV, saliency maps) require significant manual intervention, which hinders their large-scale deployability.
    In this paper, we propose a \robustness-assessment framework, at the core of which is the idea of using machine-checkable concepts.
    Our framework defines a large number of concepts that the DNN explanations could be based on and performs the explanation-conformity check at test time to assess prediction \robustness. Both steps are executed in an automated manner without requiring any human intervention and are easily scaled to datasets with a very large number of classes.
    Experiments on real-world datasets and human surveys show that our framework is able to enhance prediction \robustness significantly: the predictions marked to be \robust by our framework have significantly higher accuracy and are more robust to adversarial perturbations.
\end{abstract}

\section{Introduction}\label{sec:intro}

Explainability has emerged as an important requirement for deep neural networks (DNNs).
Explanations target a number of secondary objectives of model design (in addition to the primary objective of maximizing prediction accuracy), such as informativeness, transferability and audit of ethical values~\cite{doshikim2017Interpretability,lipton2018mythos,miller2019explanation}.
One of the most important desiderata of explainability is model \textit{\robustness}, whereby
explanations
are used to assess the extent to which some downstream task could rely on the model's predictions. For instance, a prediction classifying an input as a wolf with the explanation that the background contains snow is unlikely to be trusted by the downstream system~\cite{lime16}. A long line of research has focused on rendering DNN predictions explainable with the---often implicit---goal of assessing prediction \robustness~\cite{kim2017interpretability,lime16,chen2018looks,kim2018textual,simonyan2013deep,koh2017understanding,ross2017right,lundberg2017unified, kindermans2017learning, bahrens2010how}.

However, the scalability of these explanation-based \robustness assessment schemes is limited by the need for "humans-in-the-loop".
Prediction \robustness checks based on explanations operate as following: Given an input, one or more human-interpretable concepts are identified that have a significant impact on the model prediction. Then an \textit{explanation-conformity check} is performed to see whether the concept--prediction relationship matches human-reasoning. In the above example of wolf and snow~\cite{lime16}, a human may deem the concept--prediction relationship (snow--wolf) to be unreasonable, and consider the prediction to be \nonrobust.
However, identifying human-interpretable concepts and checking for human-reasoning requires significant human effort by the way of manual annotation of either the inputs (\eg, TCAV~\cite{kim2017interpretability}),  intermediate model components (\eg, LIME~\cite{lime16}) or both (\eg, saliency maps~\cite{simonyan2013deep}). In practice, human involvement makes many explanation-based robustness assessments unsuitable for large-scale deployment.  

\xhdr{Goals and contributions.}
In this paper, our goal is to design a highly scalable \robustness assessment framework that \textit{automates the end-to-end process of performing explanation-conformity checks}.
At the foundation of our framework are concepts with the following key properties: 
\begin{enumerate}
    \item The concepts are \itbf{identified automatically} from the training data without any  human effort.
    \item They are \itbf{machine-checkable}, \ie, they lend themselves to `concept--class' style automated explanation-conformity checks without any human involvement.
    \item They can be \itbf{added to off-the-shelf, pretrained DNNs} in a post-hoc manner to assess prediction \robustness.
\end{enumerate}

We devise an intuitive procedure for identifying machine-checkable concepts (\itbf{\concepts}) that satisfy the above key properties. Specifically, our framework automatically defines a large number of \concepts, each corresponding to features shared by some subset of one or more classes (and not shared by other classes) 
in the training data. At the end of the concept-identification process, each class in the training data has a unique set of corresponding \concepts. 
Finally, with each prediction of the DNN, our framework performs an automated explanation-conformity check to see if the \concepts corresponding to the predicted class are also detected in the learnt representations of the input (and the \concepts not corresponding to the predicted class are not detected). The predictions passing the explanation-conformity check are deemed \robust, even if individual \concepts are hard for humans to recognize.

Experiments and human surveys on real-world image classification datasets show that \concepts help increase the prediction robustness significantly. 
Specifically, we find that (i) explanation-conformant predictions are not only significantly more accurate, but their corresponding images are also easier for humans to classify confidently than non-conformant predictions, (ii) adversarial attacks against explanation-conformant predictions are significantly harder and in many cases impractical, and (iii) \concepts also provide insights into the potential causes for prediction errors.

\section{Methodology}\label{sec:methodology}

In this section, we describe our framework for \robust prediction.

\xhdr{Formal problem setup and notation.}
Let $\Dcal = \{(\xb_i, y_i)\}_{i=1}^{N}$ denote a training dataset of $N$ examples with $\xb \in \Xcal = \RR^d$ and $y\in \Ycal = \{1, 2, \ldots, K\}$.
The learning task involves obtaining a mapping $F_{\text{clf}}: \Xcal \to \Ycal$ .
For a (deep) neural network with $L$ hidden layers, this mapping consists of applying a set of parameterized layers $f_l(\xb_l, \thetab_l)$. Here, $\xb_l$ and $\thetab_l$ denote, respectively, the input and parameters of the $l^{th}$ layer.
The whole neural network mapping can be expressed as:
$F_{\text{clf}}(\xb) = f_{\text{clf}}( f_L(f_{L-1}( \ldots, f_1(\xb, \thetab_1)  )) )$, 
where the output of $f_{\text{clf}}$---or the classification layer---is a K-dimensional vector consisting of (potentially un-calibrated) probabilities,  generally obtained by applying the softmax function within the layer $f_{\text{clf}}$.
One then obtains the prediction $\hat{y} = \text{argmax} \ F_{\text{clf}}(\xb) $.
The learning then boils down to minimizing the discrepancy between the predicted and the ground-truth labels. 
For the sake of computational tractability, this discrepancy is often expressed via the (categorical) cross-entropy loss function, denoted henceforth as $\Lcal_{\text{clf}}( F_{\text{clf}}(\xb), y  ) $.

\begin{figure}[t]
    \centering
    \includegraphics[trim={0cm 0cm 0cm 0cm},clip,width=1\textwidth]{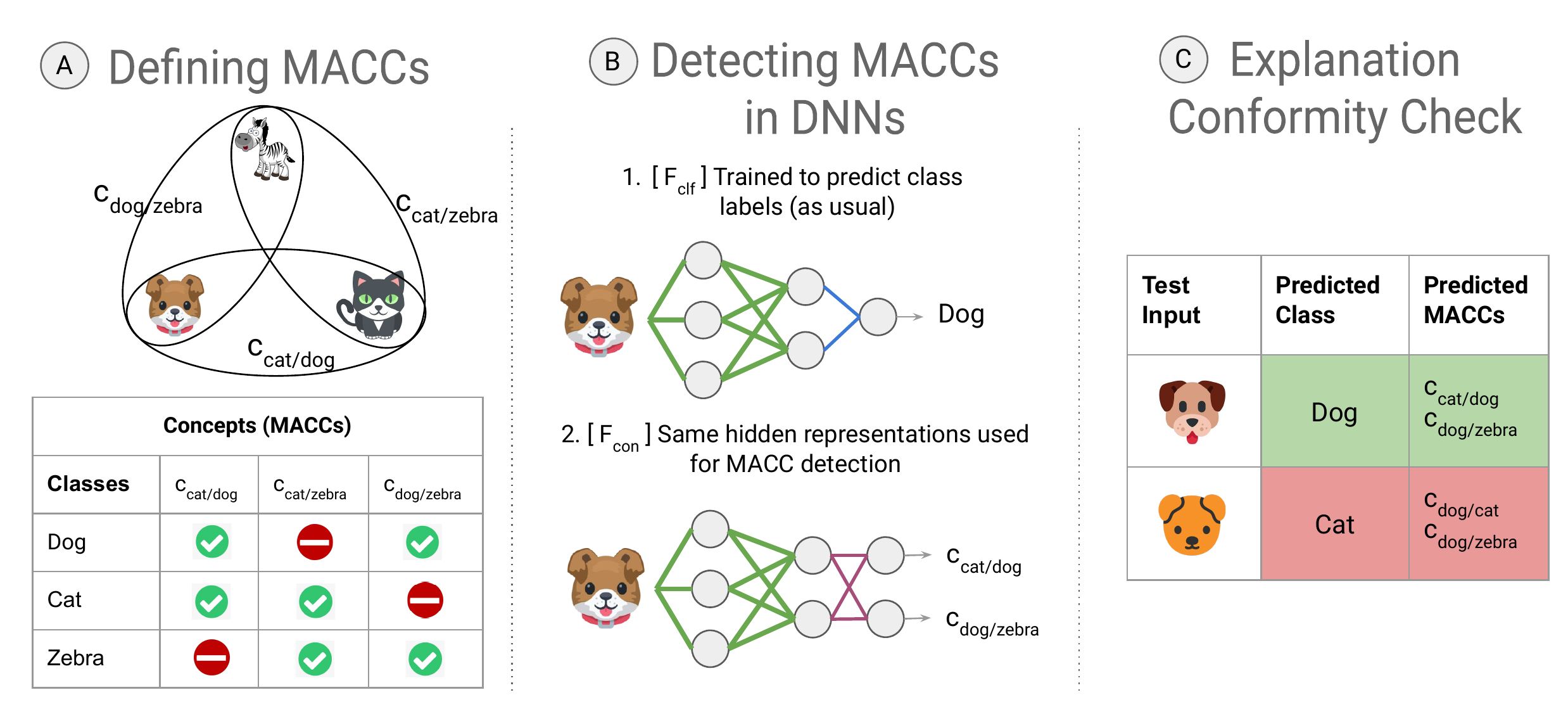}
\caption{\textbf{[System overview]} We propose the use of (a) Machine-checkable concepts (\concepts), that are defined as concepts shared between inputs of one or more classes
(Section~\ref{sec:defining_concepts})---the figure only shows the \concepts shared between two classes.
(b) 
Automatically detecting \concepts involves adding an additional classification layer (to any hidden layer) of an existing DNN (Section~\ref{sec:training_concepts}).
(c) At test time, we perform the explanation-conformity check to ensure that the \concepts corresponding to the predicted class are also detected in the image (Section~\ref{sec:prediction_with_concepts}). The predictions not passing the explanation-conformity check are deemed \nonrobust.
} 
\label{fig:sysdesign}
\end{figure}

\subsection{Our framework: \Robustness via machine-checkable concepts}

Our framework, summarized in Figure~\ref{fig:sysdesign}, consists of three main components: Defining machine-checkable concepts (\concepts), leveraging the DNN to detect \concepts, and performing explanation-conformity checks with \concepts to assess prediction \robustness. We now describe each of the components individually.

\subsubsection{Automatically defining \concepts} \label{sec:defining_concepts}

The first component of our framework  automatically defines  \concepts that are amenable to  explanation-conformity checks without any human intervention. To define \concepts, we leverage the following key insight~\cite{miller2019explanation,hesslow1988problem}: one method of composing explanations is to point to presence or absence of  \textit{concepts} in the input, where a concept is a feature that is possessed by inputs of a certain set of classes in the dataset, and not possessed by other classes.
For instance, in an animal classification task involving zebras, cats and dogs, zebras might have a unique concept \textit{stripes}~\cite{kim2017interpretability}, that is not shared by any other class. 
Similarly, dogs and cats might share a concept \textit{paws} that is not shared by any other class.

Most prior works detect these concepts by manually annotating (parts of) inputs that contain them (\eg, ~\cite{kim2017interpretability,ross2017right,lime16}).
Instead of manually annotating the inputs, \textit{for every possible subset of one or more classes}, we define one \concept that corresponds to the features shared by inputs in that subset. This way of defining \concepts leads to $M = 2^{K} -1$ concepts in a dataset with $K$ classes.
For instance, in a datasets with classes  cat, dog and zebra, one can define $2^3 -1 = 7$ \concepts, as follows \{$c_{\text{cat}}$, $c_{\text{dog}}$, $c_{\text{zebra}}$, $c_{\text{cat/dog}}$, $c_{\text{cat/zebra}}$, $c_{\text{dog/zebra}}$, $c_{\text{cat/dog/zebra}}$\}.
Figure~\ref{fig:sysdesign} shows all overlaps involving two classes. In the figure, the concept $c_{\text{dog/cat}}$ denotes a property shared by \textit{dog} and \textit{cat}, but not by \textit{zebras}. Similarly, $c_{\text{dog/zebra}}$ denotes a property possessed by \textit{zebras} and \textit{dogs} but not by \textit{cats}.

\subsubsection{Detecting \concepts in DNNs}  \label{sec:training_concepts}

Given a DNN $F_{\text{clf}}$ as in the formal setup,
trained to predict the class labels, we express the \conceptlabel detector $F_{\rsn}$ as:
$F_{\rsn}(\xb) = f_{\rsn}( f_L(f_{L-1}( \ldots, f_1(\xb, \thetab_1)  )) )$,%
\footnote{Note that $f_{\rsn}$ can be attached to any intermediate layer between $f_1$ to $f_L$.}
where the output of $f_{\rsn}$ is an M-dimensional vector consisting of (potentially un-calibrated) probabilities,  $p(\cb_i = 1 | \xb)$.
Since $F_{\rsn}$ attempts a multilabel classification task,
we obtain the probabilities using the sigmoid function  $\sigma(z) = (1 + e^{-z} )^{-1}$. 
Finally, one obtains a predicted \conceptlabel vector $\hat{\cb} = [\hat{c}_1, \ldots, \hat{c}_M]$ with $\hat{c}_i = 1$ indicating the predicted presence/absence of each \concept in the input. Here,  $F_{\rsn}(\xb)_i > 0.5$, else $\hat{c}_i = 0$. 
Learning $F_{\rsn}$ can be done by optimizing the sum of $M$ individual binary cross-entropy loss functions, with one loss function for each \conceptlabel.
We refer to this sum of loss functions as $\Lcal_{\rsn}$.

Our framework allows for the flexibility to be trained in two different ways: (1) \textbf{Post-hoc training:} Taking a pretrained DNN $F_{\text{clf}}$ as described in the formal setup, and training the \concept detection layer, $f_{\rsn}$ by attaching it to one of the hidden layers of $F_{\text{clf}}$. With this method, the pre-learnt representations of $F_{\text{clf}}$ are used and \textit{only} the parameters of $f_{\rsn}$ are learnt. (2) \textbf{Joint training:} Training all the parameters of the network from scratch, that is, training the hidden layers $f_i, \ \forall_i \in \{1 \ldots L\}$, the class label layer   $f_{\text{clf}}$, and the \conceptlabel layer $f_{\rsn}$ by minimizing the joint loss $\lambda \Lcal_{\text{clf}} + (1-\lambda) \Lcal_{\rsn}$. Here the parameter $\lambda$ trades-off the accuracy between the class labels prediction accuracy and the \conceptlabel detection accuracy, and can be determined via cross-validation.
Finally, a combination of these two techniques (\eg, selectively training only some hidden layers) can also be used.

\subsubsection{Explanation-conformity checks with \concepts} \label{sec:prediction_with_concepts}

The final component of our framework constitutes of performing an explanation-conformity check with \concepts to assess prediction \robustness. 
Our intuition is that predictions passing the check would be more robust.
Our explanation-conformity check proceeds as follows:
Given an input instance $\xb$, let $\hat{y} = F_{\text{clf}}(\xb)$ be the class prediction and $\hat{\cb} = F_{\rsn}(\xb)$ be the \conceptlabel prediction.
Then, the explanation-conformity check probes if the \concepts corresponding to the predicted class are also detected (and the \concepts not related to the predicted class are not detected). The prediction is deemed \robust if, $\frac{\sum_{i} \II[ \hat{\cb_i} = \cb^{\hat{y}}_i ]} { M } \geq t_{\rsn}$,
for some $t_{\rsn} \in [0,1]$. A higher value of $t_{\rsn}$ means that fewer predictions would pass the explanation-conformity check, however, the degree of robustness for these predictions is expected to be higher (see Section~\ref{sec:eval} for details).

\subsection{Discussion: Salient properties of \concepts} \label{sec:concept_discussion}

\xhdr{\concepts and Human-interpretability.}
Most of the existing works on concept-centered explanation-conformity checks~\cite{kim2017interpretability,lime16} use human supervision to annotate images as containing a certain concept. 
Such concepts often correspond to features that are (i) shared by certain classes and not shared by other classes in the data, and, (ii) can be easily recognized and named by humans (\eg, stripes on zebras, paws on cats and dogs).
While \concepts are not explicitly human recognizable,%
\footnote{Instead, \concepts may represent complex \textit{polymorphic and composite} features in practice, i.e., the \concept corresponding to `features shared by cats and dogs but not zebras' could correspond to a paw or the non-existence of stripes, or any combination of such distinguishing features.}
and hence do not satisfy criterion (ii), their definition procedure (Section~\ref{sec:defining_concepts}) ensures that they do indeed satisfy criterion (i).
In this sense, \concepts subsume the concepts defined in prior work on concept-centered explainability.
However, our framework trades-off human recognizability of \concepts to enable end-to-end automation of robustness assessments from \concept definition $\to$ detection $\to$ explanation-conformity checks.

\xhdr{Pruning \concepts.}
It is quite possible that in a $K$-class classification task, some classes may not share any meaningful features, and their corresponding \concepts, may not correspond to any useful concepts. For instance, the class \textit{cat} may not share any similarities with class \textit{kite}, and hence, the corresponding \concept might be meaningless. We expect these \concepts to have low detection accuracy. During the training procedure (Section~\ref{sec:training_concepts}), such \concepts 
can be dropped.
Moreover, since the possible space of \concepts is very large (for a dataset of $K=100$ classes, there are a total of $2^{100} -1 \approx 10^{30}$) possible \concepts,  one could use a random subset of \concepts, or only consider \concepts that represent properties shared by exactly two, or exactly three classes (\eg, in Figure~\ref{fig:sysdesign}).
Finally, \concepts that uniquely correspond to a class may be redundant in conformity checks and can be safely pruned.

\section{Evaluation of the \robustness framework}\label{sec:eval}

In this section, we conduct experiments and human surveys on real-world datasets to evaluate the effectiveness of our \concept framework.
Specifically, we ask whether the predictions passing the \concept explanation-conformity achieve better \robustness.

\xhdr{Evaluation metrics.} Inspired by usage of explanation-conformity checks in practice~\cite{doshikim2017Interpretability,bhatt2020explainable,tao2018attacks}, we use the following evaluation metrics to quantify prediction robustness: (i) \itbf{Error Estimability}, \ie, accuracy on explanation-conformant predictions, (ii) \itbf{Error Vulnerability}, \ie, resistance to adversarial attacks on explanation-conformant predictions, and, (iii), \itbf{Error Explainability}, \ie, ability to map errors to potential issues in the input.

\xhdr{Setup.}
We conduct experiments on CIFAR-10, CIFAR-100 and Fashion MNIST datasets. We define \concepts such that each class in CIFAR-10 and Fashion MNIST data is accompanied by 9 \concepts whereas in CIFAR-100 data, this number is 99.

We use simple deep CNN architectures, that have publicly available implementations, and provide comparable performance to state-of-the-art. Additional details on data preprocessing, \concept definition, picking $t_{\text{con}}$, and training architectures can be found in Appendix~\ref{sec:appendix_implementation_details}.

Training the models to maximize the classification accuracy leads to a test set accuracy of $88.8\%$, $92.49\%$ and $59.41\%$ on CIFAR-10, Fashion MNIST and CIFAR-100 datasets, respectively. We refer to this model as the \itbf{vanilla} model. 
For the training of $F_{\rsn}$, we consider the post-hoc training alternative considered in Section~\ref{sec:methodology}. The joint training alternative leads to similar statistics.
For the detailed analysis, we focus on the performance of post-hoc training and leave detailed comparison between different training schemes for a future study.
For \textit{performance comparison}, we use the probability calibration method of \citet{guo2017calibration} (see Section~\ref{sec:eval_discussion}).

We now present the performance of \concepts in improving prediction \robustness.

\setlength{\tabcolsep}{0.5em}
\begin{table*}[t]
\centering
\caption{\textbf{[Error Estimability]}
Accuracy of the vanilla DNN with no explanation-conformity check (Vanilla), accuracy on samples passing the explanation-conformity check (explanation-conf.) and on samples not passing the check (non explanation-conf.). 
Numbers in parentheses show the fraction of samples in each category.
Accuracy on explanation-conformant predictions is significantly higher.
}
\begin{tabular}{lccc}
\hline

 & \multicolumn{1}{c}{\textbf{\begin{tabular}[c]{@{}l@{}}Vanilla\end{tabular}}}
 & \multicolumn{1}{c}{\textbf{\begin{tabular}[c]{@{}l@{}}Explanation-conf.\end{tabular}}}
 & \multicolumn{1}{c}{\textbf{\begin{tabular}[c]{@{}l@{}}Non explanation-conf.\end{tabular}}} \\ \hline
\multicolumn{1}{l||}{\textbf{CIFAR-10}}
& \multicolumn{1}{c}{0.89 (1.00)} 
& \multicolumn{1}{c}{0.93 (0.91)}
& \multicolumn{1}{c}{0.48 (0.09)} \\
\hline
\multicolumn{1}{l||}{\textbf{Fashion-MNIST}}
&\multicolumn{1}{c}{0.92 (1.00)}
& \multicolumn{1}{c}{0.99 (0.70)}
& \multicolumn{1}{c}{0.77 (0.30)} \\
\hline
\multicolumn{1}{l||}{\begin{tabular}[c]{@{}l@{}}\textbf{CIFAR-100}\end{tabular}} 
& \multicolumn{1}{c}{0.59 (1.00)}
& \multicolumn{1}{c}{0.65 (0.84)}
& \multicolumn{1}{c}{0.30 (0.16)} \\
\bottomrule
\end{tabular}
\label{table:error_estimation}
\end{table*}

\subsection{Do \concepts provide reliable Error Estimability?} \label{sec:res_error_est}

We propose and test two hypotheses related to reliable error estimability: (i) predictions that pass the \concept explanation-conformity check are more likely to be accurate, and, (ii) predictions that are not explanation-conformant might consist of inputs with high aleatoric uncertainty~\cite{der2009aleatory} and might be more difficult for even humans to classify.

Table~\ref{table:error_estimation} shows that on all three datasets, the prediction accuracy on explanation-conformant predictions is significantly higher than non-conformant predictions, validating our hypothesis (i).
To confirm our hypothesis (ii), we show images from CIFAR-10 data to human annotators at Amazon Mechanical Turk (AMT). The AMT annotators are shown an image and asked to choose the class that the image belongs to from the list of 10 classes. Each image is annotated by 30 users. Further details on the experiment can be found in Appendix~\ref{sec:appendix_error_explainability}.

The results show that for explanation-conformant images, humans are able to detect the correct class $91.25\%$ of the time, whereas accuracy for non-conformant images is $83.19\%$. Moreover, the worker disagreement---as measured via average Shannon Entropy---is $0.22$ and $0.39$ for explanation-conformant, and non-conformant images. The difference in accuracy and worker agreement shows that the non explanation-conformant images are harder not only for the DNN, but also human annotators to classify. We expand on the difficulty of human annotators in Section~\ref{sec:error_expl}.

\subsection{Do \concepts defend against Error Vulnerability?}\label{sec:error_vulnerability}

We now ask if an explanation-conformity check can help defend against adversarial perturbations. 
Specifically, we start off with a $50\%$ random subset of test images that were correctly classified by the vanilla DNN and adversarially perturb them w.r.t. $F_{\text{clf}}$ so that they are now incorrectly classified. We use a number of popular adversarial attacks (see Table~\ref{table:adv_defences}).
Next, we check if these adversarial perturbation designed to change the class labels also resulted in a corresponding change in the detected \concepts. If that is not the case, then \concept explanation-conformity check could be used as a method to detect adversarial perturbations.

Table~\ref{table:adv_defences} shows the fraction of adversarially attacked inputs that fails the \concept explanation-conformity check, revealing that the check is able to detect a vast fraction of adversarial attacks.

While \concepts are able to defend against a significant proportion of attacks on class labels, a determined adversary could \textit{additionally} attack the \concept detection component ($F_{\text{con}}$ in Section~\ref{sec:methodology})  such that not only does the class label get switched, the \concept prediction is also changed such that the explanation-conformity check is passed.
We now study the nature of such adversarial perturbations.
To perform this attack, we modify the PGD attack (details in Appendix~\ref{sec:appendix_consistent_attack}) such that the class labels and \concepts are changed in a consistent manner to pass the explanation-conformity check.

\setlength{\tabcolsep}{0.5em}
\begin{table*}[t]
\centering
\caption{\textbf{[Error Vulnerability]} \textit{Attacking class labels}. Fraction of adversarially perturbed inputs that fail the explanation-conformity check (meaning the adversarial attack is detected.) On CIFAR-10 and Fashion-MNIST data,  $>98\%$ of the attacks are detected, except for DeepFool on CIFAR-10 where  around $40\%$ are detected. On CIFAR-100 data, around half of the adversarial attacks are detected.}
\begin{tabular}{lcccc}
\hline
 & \multicolumn{1}{c}{\textbf{\begin{tabular}[c]{@{}l@{}}FGSM~\citep{fgsm_paper}\end{tabular}}}
 & \multicolumn{1}{c}{\textbf{\begin{tabular}[c]{@{}l@{}}DeepFool~\citep{moosavi2016deepfool}\end{tabular}}}
 & \multicolumn{1}{c}{\textbf{\begin{tabular}[c]{@{}l@{}}C\&W (L2)~\citep{carliniwagner}\end{tabular}}}
 & \multicolumn{1}{c}{\textbf{\begin{tabular}[c]{@{}l@{}}PGD~\citep{aleks2017pgd}\end{tabular}}} \\ \hline
\multicolumn{1}{l||}{\textbf{CIFAR-10}}
& \multicolumn{1}{c}{0.98} 
& \multicolumn{1}{c}{0.41} 
& \multicolumn{1}{c}{1.00} 
& \multicolumn{1}{c}{0.99} \\
\hline
\multicolumn{1}{l||}{\textbf{Fashion-MNIST}}
&\multicolumn{1}{c}{1.00}
& \multicolumn{1}{c}{0.99}
& \multicolumn{1}{c}{1.00}
& \multicolumn{1}{c}{1.00} \\
\hline
\multicolumn{1}{l||}{\begin{tabular}[c]{@{}l@{}}\textbf{CIFAR-100}\end{tabular}}
& \multicolumn{1}{c}{0.50}
& \multicolumn{1}{c}{0.45}
& \multicolumn{1}{c}{0.49}
& \multicolumn{1}{c}{0.50} \\
\bottomrule
\end{tabular}
\label{table:adv_defences}
\end{table*}

\begin{figure}[t]
    \centering
    \begin{subfigure}[b]{0.30\textwidth}
        \includegraphics[trim={9cm 3cm 9cm 3cm},clip,width=1\textwidth]{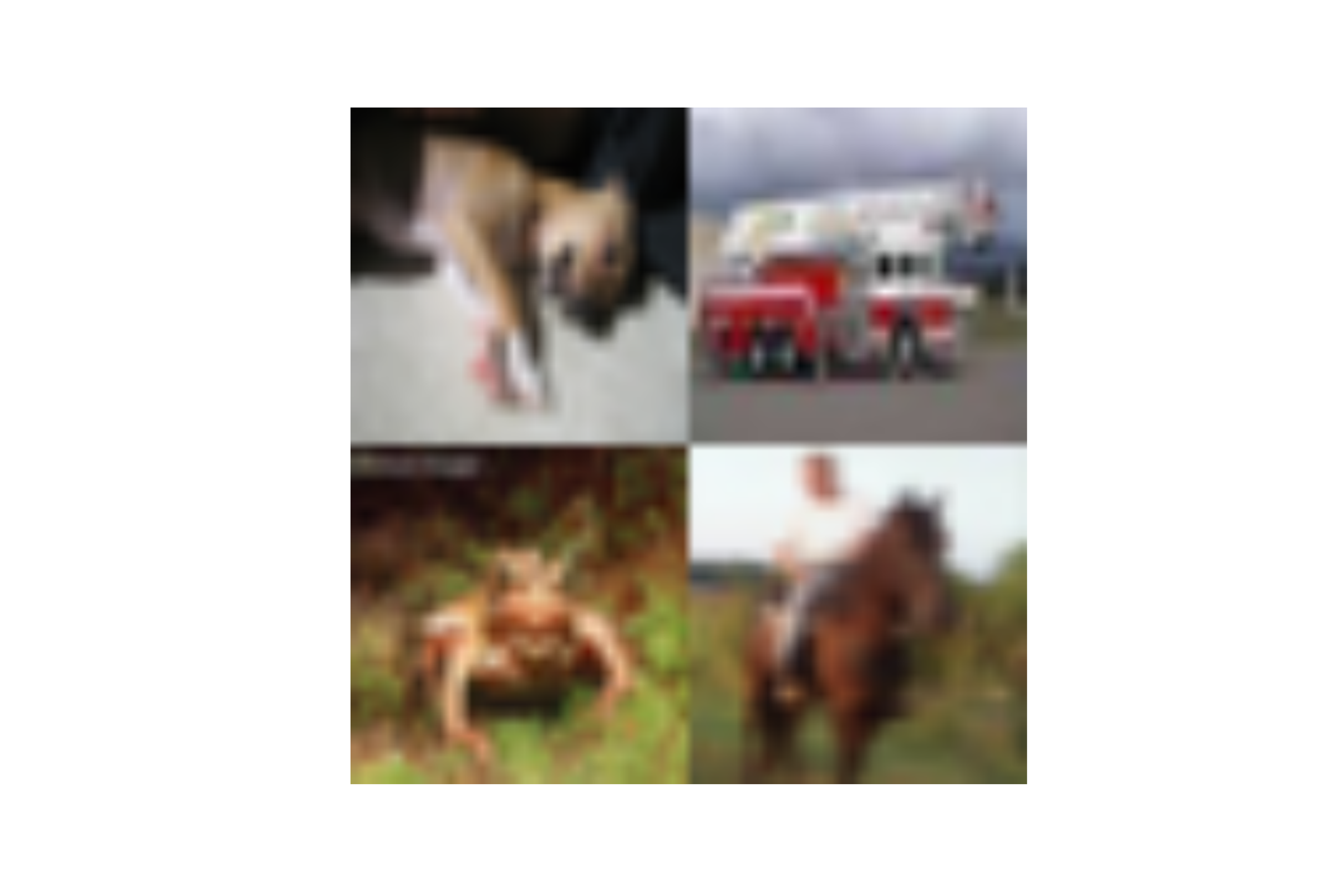}
        \caption{Original Images}
        \label{fig:original_cifar10}
    \end{subfigure}
    \begin{subfigure}[b]{0.30\textwidth}
        \includegraphics[trim={9cm 3cm 9cm 3cm},clip,width=1\textwidth]{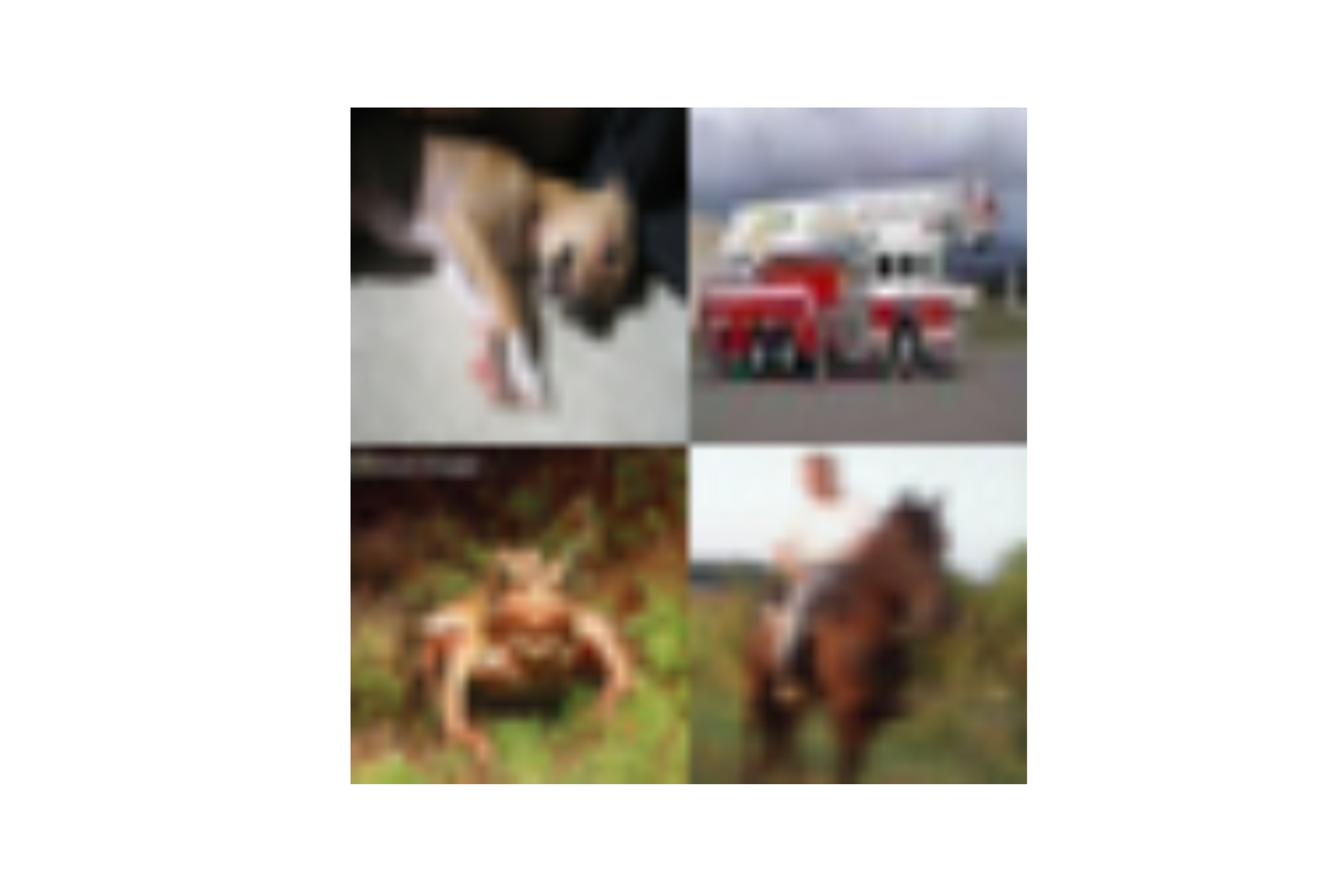}
        \caption{PGD attack class-only}
        \label{fig:pgd_cifar10}
    \end{subfigure}
    \begin{subfigure}[b]{0.30\textwidth}
        \includegraphics[trim={9cm 3cm 9cm 3cm},clip,width=1\textwidth]{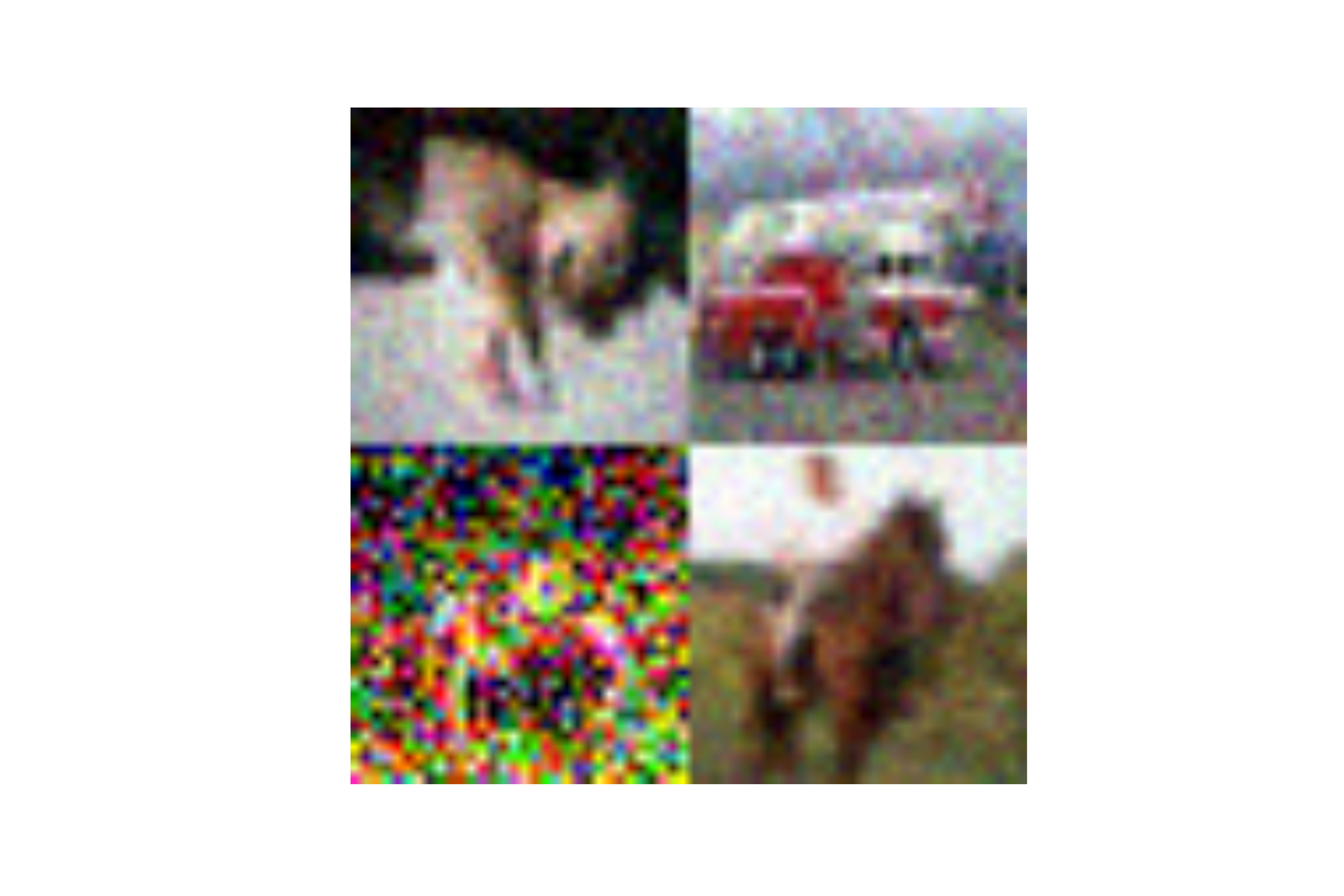}
        \caption{PGD attack explanation-conf.}
        \label{fig:pgd_consistent_cifar10}
    \end{subfigure}
\caption{ \textbf{[Error Vulnerability]}
\textit{Attacking both class labels and \concepts.}
The figure shows some randomly selected \textit{Original Images} from CIFAR-10 that were correctly classified by the vanilla DNN.
Also shown are the perturbed images obtained by performing a conventional adversarial attack, using the PGD method~\cite{aleks2017pgd}, aimed at switching the predicted class label on the input (\textit{Class-only}). As is expected of such attacks~\cite{papernot2017practical,carlini2017towards}, the perturbations are imperceptible to the human eye.
Finally, we show the perturbed images where the adversarial attack not only changes the predicted class labels, but also the \concepts such that the predictions are explanation-conformant (\textit{Explanation-conf.}).
Explanation-conformant perturbations are so large that they are clearly perceptible to the human eye. 
} 
\label{fig:pgd_consistent_images}
\end{figure}

We note that the perturbation required to perform an explanation-conformant attack is significantly higher than the one required for an attack that aims to change the class label only. Specifically, while the class-only attacks in Table~\ref{table:adv_defences}) require a perturbation (based on L2 distance from the original image) of $0.31 \pm 0.20$ and $0.26 \pm 0.14$ on CIFAR-10 and Fashion-MNIST datasets respectively, the explanation-conformant perturbations have a magnitude of $5.31 \pm 5.62$ and $3.16 \pm 2.85$. In other words, \textit{explanation-conformant attacks require perturbations that are more than an order of magnitude larger}.

\xhdr{Are the perturbations still imperceptible to humans?}
We suspect that the magnitude of the explanation-conformant perturbations is so large that they might not be imperceptible to humans anymore. Perturbations being imperceptible to humans is often considered as a major property adversarial perturbations~\cite{papernot2017practical,carlini2017towards}.
To test this hypothesis, we set up a human survey on AMT where the humans are shown three kinds of images: (i) the \itbf{original}, unperturbed image, (ii) image with \itbf{class-only} perturbation that aims to change the predicted class label, and, (iii) the image with \itbf{explanation-conformant} perturbation that aims to change the predicted class label as well as predicted \concepts such that the prediction passes the explanation-conformity check. AMT workers were then asked to label if the image contained an adversarial perturbation or not. Details of the survey can be found in Appendix~\ref{sec:error_vulnerability_human_survey}.

The results show that for class-only category, humans are able to detect the images with adversarial perturbations around $49.8\%$ of the time, \ie, the human accuracy is as good as a random guess. On the other hand, for images in the explanation-conformant category, the humans are able to detect the adversarially perturbed images $85\%$ of the time. This vast difference in human detection accuracy shows that explanation-conformant perturbations are much more noticeable to the human eyes that class-only perturbations. Figure~\ref{fig:pgd_consistent_images} also shows some examples of the explanation-conformant perturbations (more examples in Appendix~\ref{sec:appendix_error_vulnerability}). In summary, the survey shows that \textit{it is difficult to attack the \concept explanation-conformity check in a manner that is undetectable by humans. }

\subsection{Do \concepts provide insights into the causes of errors?}\label{sec:error_expl}

\begin{figure}
\centering
    \begin{subfigure}[b]{0.45\textwidth}
    \begin{tabular}{l||cc}
    \hline
    \thead{Human \\ agreement} & \thead{$< 6$ \concepts \\ detected} & \thead{$\geq 6$ \concepts \\ detected} \\
    \hline
    {\shortstack{$\leq 50\%$}} & $75.4\%$ & $24.6\%$ \\
    {\shortstack{$> 50\%$}} & $47.7\%$ & $52.3\%$ \\
    \bottomrule
    \end{tabular}
    \caption{Human agreement \& num. detected \concepts}
    \label{table:plurality_share_cdf}
    \end{subfigure}
    \hspace{5mm}
    \begin{subfigure}[b]{0.5\textwidth}
        \includegraphics[width=1\textwidth]{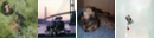}
        \caption{Images with lowest  agreement}
        \label{fig:least_plurality_images}
    \end{subfigure}
\caption{
\textbf{[Insights into causes of errors]}
The table~(\ref{table:plurality_share_cdf})  shows that images with less human agreement also tend to have few detected \concepts.
The figure~(\ref{fig:least_plurality_images}) shows the images with lowest human agreement.
For more examples and details, see Appendix~\ref{sec:appendix_error_explainability}.
} 
\label{fig:plurality_shares}
\end{figure}

Inspired by the insight in Section~\ref{sec:res_error_est} that even humans tend to make more errors on non explanation-conformant inputs, we now further explore these cases.

Specifically, we note that some non explanation-conformant inputs consists of cases where $F_{\text{con}}$ is able to detect very few \concepts (see Appendix~\ref{sec:appendix_error_explainability_results} for a full distribution).%
\footnote{An explanation-conformant prediction, with $t_{\text{con}}=1$ in CIFAR-10 data would mean that $F_{\text{con}}$ detects 9 exactly \concepts in the input. See Appendix~\ref{sec:appendix_implementation_details} for details on \concepts for each class.}
This means that the DNN is struggling to identify concepts related to \textit{any} class in the input. We hypothesize that low concept detection rate might mean that these inputs might consist of cases where even humans might find it hard to identify the class of the image.

To test this hypothesis, we  divide the non explanation-conformant images from the annotation task described in Section~\ref{sec:res_error_est} into different categories based on the (dis)agreement between human annotators. The agreement here is measured as the fraction of the votes obtained by the class with most votes. Hence, an agreement value of $1.0$ means that all humans annotated the image with the same class, whereas a value of $0.1$ means that the most-voted-for class received votes that are no better than a random assignment (as the CIFAR-10 dataset consists of 10 classes).

Next, we divide the images into two categories: images where $< 6$ \concepts were detected and where $\geq 6$ \concepts were detected. 
Figure~\ref{table:plurality_share_cdf} shows the relative fraction of these two categories against the human agreement.
The figure shows that the images with small degree of agreement tend to mostly consist of cases where very few ($<6$ \concepts) are detected.
Specifically, out of the images with less or equal to $50\%$ agreement, $75.4\%$ of them have $5$ or less \concepts detected.
Figure~\ref{fig:least_plurality_images} shows the images with lowest human agreement.

These results show that detection of very few \concepts in an image correlates with the fact that even human judges (who are often the source of ground truth in image classification tasks) would  find it difficult to classify these images. Hence, \textit{\concepts can serve as a useful tool to pinpoint problematic inputs in the data}. However, we do note that \concepts are not able to explain causes of errors for \textit{all} the misclassified inputs, rather they only explain errors for a certain category of the data (with very few concepts detected).

\subsection{Discussion} \label{sec:eval_discussion}

The results show that \concepts can be used to perform explainability checks that significantly enhance predictions' \robustness along a wide range of measures. In this section, we discuss some more pertinent points related to the implementation of \concepts.

\xhdr{Effect of varying $t_{\text{con}}$.}
As described in Section~\ref{sec:methodology}, varying $t_{\text{con}}$ can be thought of as a flexible parameter to fine-tune prediction \robustness.
We further investigate the effect of $t_{\text{con}}$ on the fraction of samples deemed explanation-conformant and the prediction accuracy on these samples. Results in Appendix~\ref{sec:appendix_error_estimability} shows that increasing $t_{\text{con}}$ leads to more samples being marked as explanation-conformant, however, the classification accuracy on explanation-conformant samples decreases.

\xhdr{Other methods for assessing prediction \robustness.}
We also compare the \robustness estimates obtained using the \concept explanation-conformity check with the more traditional method of probability calibration.
Specifically, we use the temperature scaling method of~\citet{guo2017calibration} to calibrate the softmax probabilities.%
\footnote{We use the implementation provided by the authors: \small{\url{github.com/gpleiss/temperature_scaling}}}
We then predictions to be \robust  if the (calibrated) predicted class probability is above $X$, where $X$ is chosen such that the same fraction of predictions are marked robust as by our method in Table~\ref{table:error_estimation}.%
\footnote{Comparison with more thresholds reveals similar insights. Details in Appendix~\ref{sec:appendix_error_estimability}.}

The comparison reveals that (i) both the robustness check based on calibrated probabilities and the \concept explanation-conformity check achieve comparable performance in terms of  the tradeoff between predictions marked \robust and classification accuracy on these predictions,  however, (ii) the calibration method leads to much lower performance in terms of Error Vulnerability, \ie, the amount of perturbation required to pass the calibration \robustness check is almost an order of magnitude smaller. More details on the comparison can be found in Appendix~\ref{sec:appendix_error_vulnerability}. 

\section{Related work} \label{sec:related}

Most prior approaches to DNN explainability and \robustness operate by identifying important features, concepts, or training data instances~\cite{lime16, chen2018looks, gulshad2019interpreting,NIPS2018_8231,simonyan2013deep,koh2017understanding, khanna2018interpreting,kim2016examples,bien2011prototype,doshikim2017Interpretability}.
The main differences between  these studies and our approach is that we target a specific application of concept explainability, \ie, the explanation-conformity check, and automate the robustness assessment procedure.

A line of work closely related to ours is that of concept-based explanations. 
\citet{kim2017interpretability} propose a method to evaluate how important a user-defined concept is in predicting a specific class.
\citet{yeh2019completenessaware} propose ways to find concepts that are enough to explain a given prediction.
\citet{ghorbani2019towards} proposed ways to automatically extract concepts from visual data while \citet{bouchacourt2019educe} proposed a similar approach for textual data.
\citet{goyal2019explaining, shi2020conceptbased} focus on identifying human-interpretable concepts that have causal relationships with model's predictions.
However,  none of these methods proposes automation of explanation-conformity checks.

Some recent studies~~\cite{tao2018attacks,ghorbani2017fragile} have focused on linking explainability and adversarial robustness.
~\citet{ghorbani2017fragile} show that saliency map based explanations are easy to fool via adversarial attacks. On the other hand, \conceptlabels are quite resistant to adversarial perturbations
(Section~\ref{sec:error_vulnerability}).
~\citet{tao2018attacks} propose an explanation-based check to detect adversarial perturbations. However their approach is limited to hand-crafted features and
is specialized for facial recognition,
whereas our approach can be extended to more general image recognition tasks and also other classification tasks.

Prediction \robustness has also been studied in the context of calibration and prediction uncertainty~\cite{madras2018predict,guo2017calibration,gal2015dropout,cortes2016learning,pmlr-v97-geifman19a}. 
Empirical comparison with a recent calibration technique~\cite{guo2017calibration} shows that while the robustness check based on this technique provides comparable accuracy,
\concepts are far more robust to adversarial perturbations (Section~\ref{sec:eval_discussion}), and additionally help provide insights into the causes of errors (Section~\ref{sec:error_expl}). 
Moreover, unlike many prior works in this line of research, \eg, \cite{madras2018predict,pmlr-v97-geifman19a},  our proposed framework can be easily plugged into an existing trained model in a post-hoc manner.

Finally, \concepts also share some similarities with redundant output encoding and error correcting output codes (ECOC)~\cite{berger1999error,ghani2000using}. However, unlike \concepts,  ECOCs do not provide an explanation-conformity check with a built-in reject option.

\section{Conclusion, limitations 
\& future work}
\label{sec:conclusion}

In this work, we proposed a robustness assessment framework that uses  Machine-checkable Concepts, or \concepts, to automate the end-to-end process of performing explanation-conformity checks. 
The automation means that our framework can be scaled to a large number of classes.
\concepts partly achieve this scalability by focusing on a specific explainability desideratum---\ie, assessment of prediction robustness---and potentially sacrificing some other desiderata  (details in Section~\ref{sec:intro}).
Experiments and human-surveys on several real-world datasets show that
the \concept 
explanation-conformity check facilitates higher prediction accuracy (on predictions passing the explanation-conformity check), adds resistance to adversarial perturbations, and can also help provide insights into the source of errors.

Our work opens several avenues for future work: For now, \concepts are defined such that they are shared between all images of the same class. 
A useful follow-up would be to consider \textit{multiple sets} of \conceptlabels per class to account for intra-class variability.
Moreover exploring the \concept pruning strategies, analyzing the effect of the number of \concepts on the robustness, and 
a deeper exploration of the tradeoffs provided by different training methodologies mentioned in Section~\ref{sec:methodology} (post-hoc, joint, or a combination) are also  promising future directions.

We believe that our work has potential to provide significant positive impact for the society. As machine learning models are deployed in a wide array of real-world domains, the issue of prediction robustness has become increasingly relevant. The ability of our methods to provide improved uncertainty estimates, offer resistance to adversarial perturbations, and the capability to potentially debug the model errors is a useful tool for many societal applications. Examples of these applications include image search in online databases and driver-assistance systems in the automotive domains.

On the flip side, our methods are evaluated empirically and do not come with theoretical performance guarantees. As a result, appropriate care should be applied before using them in critical life-affecting domains. An analysis exploring the performance guarantees remains an important future research direction.

Most of the prior work on concept-based explanations restricts itself to concepts that can be explicitly named by humans (see Section~\ref{sec:concept_discussion} for a discussion). Our framework represents a departure from this restriction, and places more emphasis on machine-checkability (much like the line of work on machine-checkable theorem proving~\cite{zammit1999readability}). As a result, while our machine-checkable concepts (\concepts) are able to meet the goal that they were designed for, it should be noted that they may not fulfil some other explainability criteria~\cite{doshikim2017Interpretability,bhatt2020explainable,lipton2018mythos}. Combining machine-checkability with human-interpretability would be a worthwhile future research direction.
\section{Acknowledgements}

Dickerson and Nanda were supported in part by NSF CAREER Award IIS-1846237, DARPA GARD \#HR00112020007, DARPA SI3-CMD \#S4761, DoD WHS Award \#HQ003420F0035, and a Google Faculty Research Award. This work was supported in part by an ERC Advanced Grant “Foundations for Fair Social Computing” (no. 789373). 

\bibliographystyle{plainnat}
\bibliography{main}  %
\appendix
\newpage

\section{Implementation details and reproducibility}\label{sec:appendix_implementation_details}

In this section, we report relevant implementation details from Section~\ref{sec:eval}.

\subsection{Extracting \conceptlabels}\label{sec:appendix_extracting_concepts}

As discussed in section~\ref{sec:defining_concepts}, one could define \conceptlabels as concepts shared between one or more classes. Since this quantity scales exponentially with the number of classes, in the paper, we restrict ourselves to \conceptlabels that are shared by every pair of classes.
Table~\ref{table:concept_exmaples_datasets} shows some example of \conceptlabels for each of the datasets used in Section~\ref{sec:eval} (\ie, CIFAR-10, CIFAR-100 and Fashion-MNIST).
By restricting ourselves to \conceptlabels shared by each pair of classes, we get $\Comb{N}{2}$ unique \conceptlabels, and each class has $N - 1$ unique \conceptlabels. Appendix~\ref{sec:appendix_error_estimability} provides some initial analysis on \conceptlabels defined as shared concepts between each triplet of classes, thus leading to $\Comb{N}{3}$ concepts.
However, we leave a detailed analysis of different combinations of concepts to future work.

\setlength{\tabcolsep}{1.0em}
\begin{table*}[h]
\centering
\caption{\textbf{[\conceptlabels details]}
Details and examples of \conceptlabels extracted for each dataset. \conceptlabels here correspond to concepts shared between each pair of classes.
}
{\renewcommand{\arraystretch}{1.25}
\begin{tabular}{lccc}
\hline

 & \multicolumn{1}{c}{\textbf{\begin{tabular}[c]{@{}l@{}}Total \# of\\unique \conceptlabels\end{tabular}}}
 & \multicolumn{1}{c}{\textbf{\begin{tabular}[c]{@{}l@{}}\# \conceptlabels\\per class\end{tabular}}}
 & \multicolumn{1}{c}{\textbf{\begin{tabular}[c]{@{}l@{}}Examples of\\\conceptlabels\end{tabular}}} \\ \hline
\multicolumn{1}{l||}{\textbf{CIFAR-10}}
& \multicolumn{1}{c}{45} 
& \multicolumn{1}{c}{9}
& \multicolumn{1}{c}{\begin{tabular}[c]{@{}l@{}} $c_{dog/cat}$, $c_{dog/airplane}$,\\$c_{dog/truck}$, $c_{dog/automobile}$ ... \end{tabular}} \\
\hline
\multicolumn{1}{l||}{\begin{tabular}[c]{@{}l@{}}\textbf{CIFAR-100}\end{tabular}} 
& \multicolumn{1}{c}{4095}
& \multicolumn{1}{c}{99}
& \multicolumn{1}{c}{\begin{tabular}[c]{@{}l@{}} $c_{baby/boy}$, $c_{baby/girl}$,\\$c_{baby/man}$, $c_{baby/woman}$ ... \end{tabular}} \\
\hline

\multicolumn{1}{l||}{\textbf{Fashion-MNIST}}
&\multicolumn{1}{c}{45}
& \multicolumn{1}{c}{9}
& \multicolumn{1}{c}{\begin{tabular}[c]{@{}l@{}} $c_{trouser/pullover}$, $c_{trouser/dress}$,\\$c_{trouser/coat}$, $c_{trouser/sandal}$ ... \end{tabular}} \\

\bottomrule
\end{tabular}
}
\vspace{-2mm}
\label{table:concept_exmaples_datasets}
\end{table*}

\subsection{Model Architectures}\label{sec:appendix_model_architectures}
Our goal is to use architectures that are: (i) easily implementable and are widely used in the community, and, (ii)  are able to achieve close to state-of-the-art accuracy on the respective datasets.
To this end, we use the following architectures.
\begin{itemize}
    \item \textbf{CIFAR-10: } The architecture by \citet{cifar-zagoruyko} that achieves competitive accuracy~\cite{cifar-benenson}.
    \item \textbf{Fashion-MNIST: } The architecture as described in~\cite{cifar-fmnist}, taken from the official Github repository of the dataset.\footnote{\url{https://github.com/zalandoresearch/fashion-mnist}}
    \item \textbf{CIFAR-100: } The architecture as described in~\cite{cifar100-git}.
\end{itemize}

Furthermore, for the sake of simplicity, we chose to train these models from scratch, and did not rely on transfer learning or pretrained feature extractors.  We leave the detailed analysis of \concepts under these training paradigms to a separate future study.

\subsection{Data Preprocessing}
For both CIFAR10 and Fashion-MNIST we do mean-std. normalization. For CIFAR-10 and CIFAR-100 we use a mean and std of $(0.5, 0.5, 0.5)$ while for Fashion-MNIST we use a mean and std of $0.1307$ and $0.3081$ respectively. These are commonly used values for the respective datasets. 
While all 3 datasets come with a pre-defined train-test split, we further split the train set into train and validation set with 10k samples in the validation set for each of the datasets. Table~\ref{table:train_val_test_split} shows the number of samples in train, val and test sets for each of the datasets.

\setlength{\tabcolsep}{1.0em}
\begin{table*}[t]
\centering
\caption{\textbf{[Dataset splits]}
Training, validation and test set splits for each dataset.
}
{\renewcommand{\arraystretch}{1.25}
\begin{tabular}{lccc}
\hline

 & \multicolumn{1}{c}{\textbf{\begin{tabular}[c]{@{}l@{}}Train\end{tabular}}}
 & \multicolumn{1}{c}{\textbf{\begin{tabular}[c]{@{}l@{}}Validation\end{tabular}}}
 & \multicolumn{1}{c}{\textbf{\begin{tabular}[c]{@{}l@{}}Test\end{tabular}}} \\ \hline
\multicolumn{1}{l||}{\textbf{CIFAR-10}}
& \multicolumn{1}{c}{40,000} 
& \multicolumn{1}{c}{10,000}
& \multicolumn{1}{c}{10,000} \\
\hline
\multicolumn{1}{l||}{\begin{tabular}[c]{@{}l@{}}\textbf{CIFAR-100}\end{tabular}} 
& \multicolumn{1}{c}{40,000}
& \multicolumn{1}{c}{10,000}
& \multicolumn{1}{c}{10,000} \\
\hline

\multicolumn{1}{l||}{\textbf{Fashion-MNIST}}
&\multicolumn{1}{c}{50,000}
& \multicolumn{1}{c}{10,000}
& \multicolumn{1}{c}{10,000} \\

\bottomrule
\end{tabular}
}
\vspace{-2mm}
\label{table:train_val_test_split}
\end{table*}

\subsection{Hyperparameters}
The optimizer used throughout the experiments is SGD with a momentum of $0.9$ and a learning rate of $0.01$. 

We select $t_{\text{con}}$ to be $1.0$ for CIFAR-10 and Fashion-MNIST datasets, and $0.98$ for CIFAR-100 dataset. This value is chosen to ensure that a reasonably high number of predictions are deemed explanation-conformant. A detailed trade-off between $t_{\text{con}}$ and classification accuracy is shown in Figure~\ref{fig:consistency_levels}.

\subsection{Hardware}
We use a machine with a NVIDIA Tesla p100-sxm2 16GB GPU, with 80 CPU cores and 512GB of RAM. However, all of our models and data are fairly common and can run on any standard machine without compromising the training time significantly. Depending on batch size, one can run our code in as little as 500MB of GPU space and by loading each batch from disk, RAM usage can be reduced to as little as 16GB.

\section{\conceptlabel fine-tuning \& comparison to calibration methods}\label{sec:appendix_error_estimability}

\begin{figure}[h]
    \centering
    \begin{subfigure}[b]{0.3\textwidth}
        \includegraphics[trim={0cm 0cm 0cm 0cm},clip,width=1\textwidth]{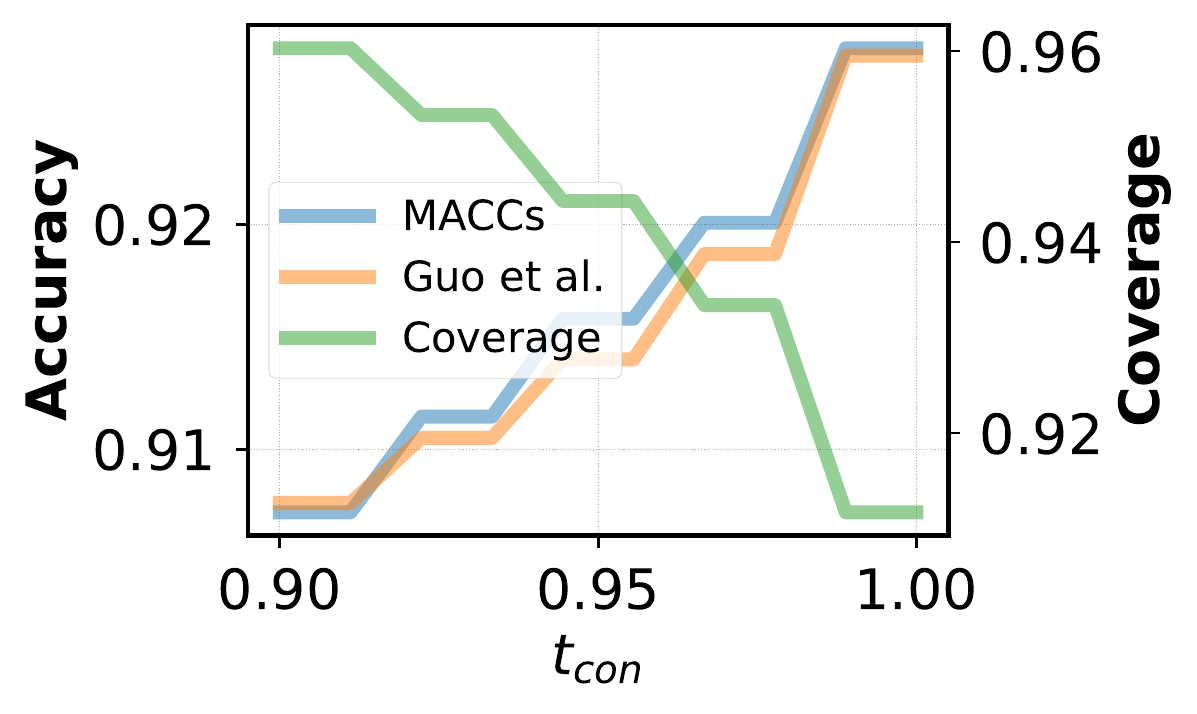}
        \caption{CIFAR-10}
        \label{fig:consistency_levels_cifar10}
    \end{subfigure}
    \begin{subfigure}[b]{0.3\textwidth}
        \includegraphics[trim={0cm 0cm 0cm 0cm},clip,width=1\textwidth]{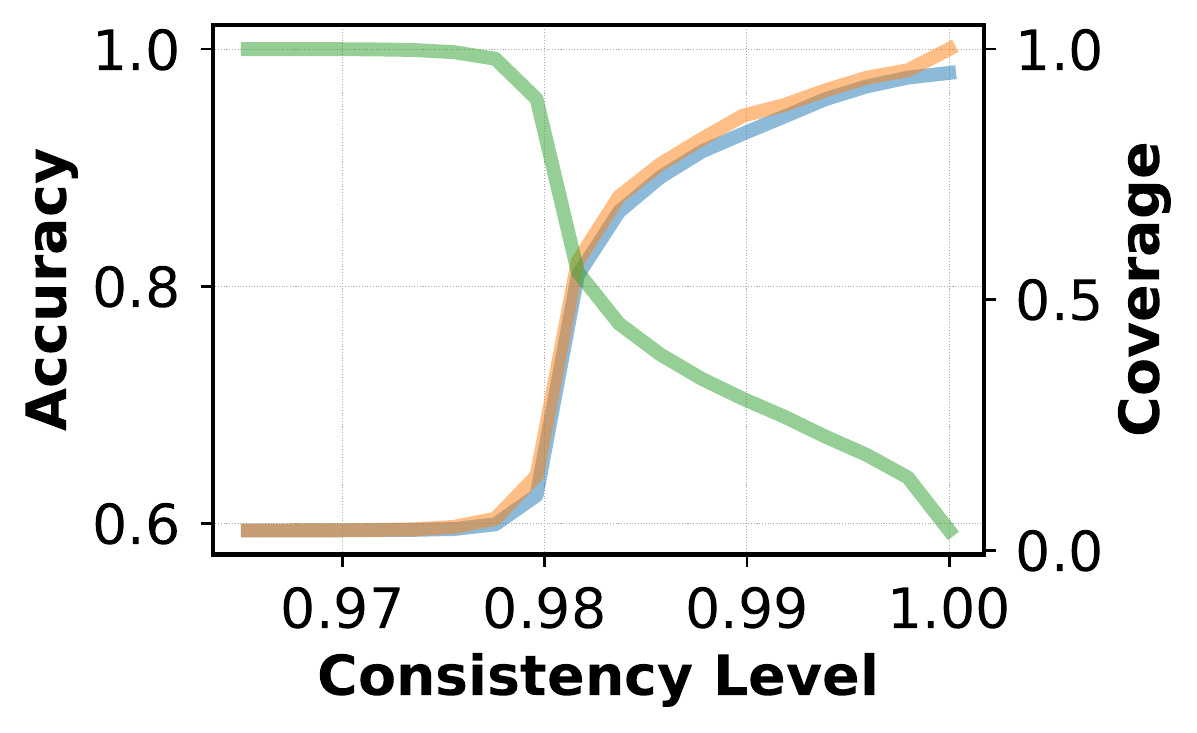}
        \caption{CIFAR-100}
        \label{fig:consistency_levels_cifar100}
    \end{subfigure}
    \begin{subfigure}[b]{0.3\textwidth}
        \includegraphics[trim={0cm 0cm 0cm 0cm},clip,width=1\textwidth]{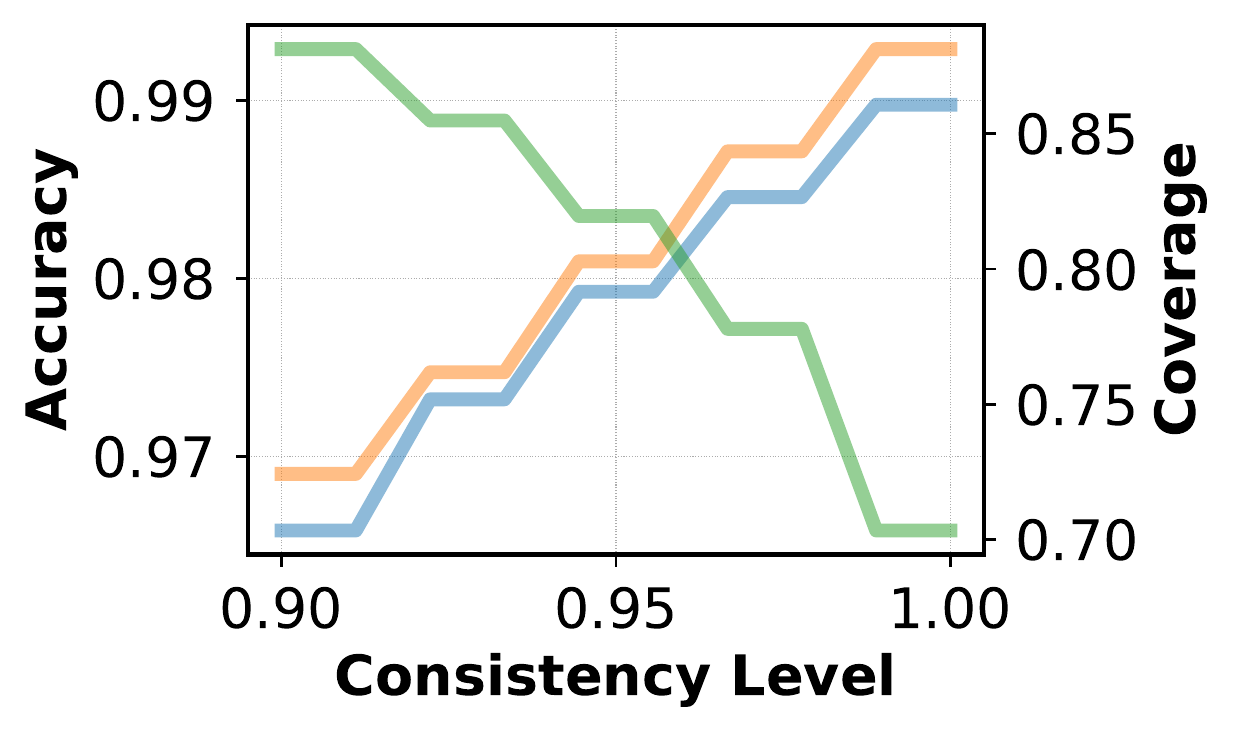}
        \caption{Fashion MNIST}
        \label{fig:consistency_levels_fashion_mnist}
    \end{subfigure}
\caption{ [Error Estimability]
A higher value of $t_{con}$ results in fewer samples being marked explanation-conformant, but the accuracy on these samples increases. Also shown is the accuracy of the probability calibration method of~\citet{guo2017calibration}. For the same level of coverge,
both methods achieve very similar accuracy.
For \concepts, coverage is the fraction of samples marked explanation-conformant. For \citet{guo2017calibration}, coverage is the fraction of samples with the calibrated prediction probability higher than the corresponding threshold (Section~\ref{sec:eval_discussion}).
} 
\label{fig:consistency_levels}
\end{figure}

\xhdr{Effect of $t_{\text{con}}$.}
Figure~\ref{fig:consistency_levels} shows the trade-off between the accuracy on explanation-conformant samples and the amount of samples which are marked explanation-conformant (coverage). As described in Section~\ref{sec:prediction_with_concepts}, and (Section~\ref{sec:eval_discussion}), this trade-off is achieved by varying  $t_{con}$. The figure shows that as expected, increasing $t_{con}$ results in fewer samples being marked explanation-conformant, however, the classification accuracy on explanation-conformant sections is higher.

\xhdr{Comparison with calibration method (\citet{guo2017calibration})}
Figure~\ref{fig:consistency_levels} also shows the trade-off between fraction of samples marked robust and accuracy on these samples achieved by using the probability calibration method of~\citet{guo2017calibration} (details in Section~\ref{sec:eval_discussion}). The figure shows that \concepts are able to achieve a trade-off very similar to that of \citet{guo2017calibration}, highlighting the competitive calibration capability of \concept explanation-conformity check.

\xhdr{Effect of number of concepts} Table~\ref{table:error_estimation_different_macs} shows the comparison on CIFAR-10 between different sets of \conceptlabels. With a higher number of \conceptlabels, we see that a slightly lower fraction of samples are marked as explanation-conformant, however, the accuracy on this set is slightly higher.
We leave a more in-depth analysis of different sets of \conceptlabels to future work.

\setlength{\tabcolsep}{0.5em}
\begin{table*}[t]
\centering
\caption{\textbf{[Error Estimability on different sets of \conceptlabels]} Performance of CIFAR-10 with different sets of \conceptlabels shows a slight trend of fewer samples being marked explanation-conformant with a higher number of \conceptlabels.
The first number is accuracy and the second number in parentheses is coverage.
}
\begin{tabular}{lccc}
\hline

 & \multicolumn{1}{c}{\textbf{\begin{tabular}[c]{@{}l@{}}Vanilla\end{tabular}}}
 & \multicolumn{1}{c}{\textbf{\begin{tabular}[c]{@{}l@{}}Explanation-conf.\end{tabular}}}
 & \multicolumn{1}{c}{\textbf{\begin{tabular}[c]{@{}l@{}}Non explanation-conf.\end{tabular}}} \\ \hline
\multicolumn{1}{l||}{\textbf{\begin{tabular}[c]{@{}l@{}}CIFAR-10 with\\$\Comb{N}{2} = 45$ \conceptlabels\\$t_{con} = 1.0$\end{tabular}}}
& \multicolumn{1}{c}{0.89 (1.00)} 
& \multicolumn{1}{c}{0.93 (0.91)}
& \multicolumn{1}{c}{0.48 (0.09)} \\
\hline
\multicolumn{1}{l||}{\textbf{\begin{tabular}[c]{@{}l@{}}CIFAR-10 with\\$\Comb{N}{3} = 120$ \conceptlabels\\$t_{con} = 1.0$\end{tabular}}}
& \multicolumn{1}{c}{0.89 (1.00)} 
& \multicolumn{1}{c}{0.94 (0.89)}
& \multicolumn{1}{c}{0.51 (0.11)} \\
\hline
\bottomrule
\end{tabular}
\label{table:error_estimation_different_macs}
\end{table*}

\section{\concepts and error interpretability}\label{sec:appendix_error_explainability}

We describe the human surveys conducted in Sections~\ref{sec:res_error_est} and~\ref{sec:error_expl}.

In the following, we describe the dataset used in the survey, the survey setup and metrics used and finally show the obtained results.

\subsection{Details of Human Experiments}

\setlength{\tabcolsep}{0.5em}
\begin{table}[t]
    \centering
    \caption{\textbf{[Survey dataset statistics]}
    Number of images from each partition used in the survey and share of the partitions in the original CIFAR-10 test set.
    }
    \begin{tabular}{c||c|c}
        \hline
        \thead{Image Group} & \thead{Number of images \\ used in the survey} & \thead{Fraction in \\ the dataset} \\
        \hline
        \thead{Correct and Explanation Conformant (C+)} & $50$ & $84.6\%$ \\
        \thead{Incorrect and Explanation Conformant (C-)} & $50$ & $6.6\%$ \\
        \thead{Less Than 6 \concepts (NC<6)} & $72$ & $0.7\%$ \\
        \thead{Between 6 and 12 \concepts (NC-6-12)} & $50$ & $6.9\%$ \\
        \thead{Greater Than 12 \concepts (NC>12)} & $122$ & $1.2\%$ \\
        \bottomrule
    \end{tabular}
    \vspace{-2mm}
    \label{tab:error_exp_data_fractions}
\end{table}

\xhdr{Dataset.}
We are interested in examining links between \concept explanation conformity checks and difficulty that humans (who are often sources of ground truth for such tasks) would face in classifying the image.
We therefore use the setup described in Appendix~\ref{sec:appendix_implementation_details} to obtain \concept- and class-predictions for the images in the CIFAR-10 test set.
Next, we divide the dataset according to explanation-conformity into five types of images:
\begin{enumerate}
    \item C+: correctly predicted by DNN and explanation-conformant
    \item C-: incorrect and explanation conformant
    \item NC<6: non- explanation-conformant with fewer than 6 predicted \concepts
    \item NC-6-12: non- explanation-conformant with between 6 and 12 predicted \concepts
    \item NC>12: non-conformant with more than 12 predicted \concepts
\end{enumerate}

From each partition we randomly sample 50 images, except for NC<6 and NC>12 which consist of very few images (72 and 122 images,  respectively), for which we therefore include all images.%
\footnote{A complete random subsampling without regard to the categories would result in near-zero images from sparsely populated categories.}
This gives us a total of 344 images used in the survey.
The composition of the set of images used in the survey as well as the fraction of images in the original CIFAR-10 test set that belong to each of the groups is shown in Table \ref{tab:error_exp_data_fractions}.

\begin{figure}[t]
    \centering
    \begin{subfigure}[b]{0.325\textwidth}
        \includegraphics[trim={4cm 1cm 4cm 0cm},clip,width=1\textwidth]{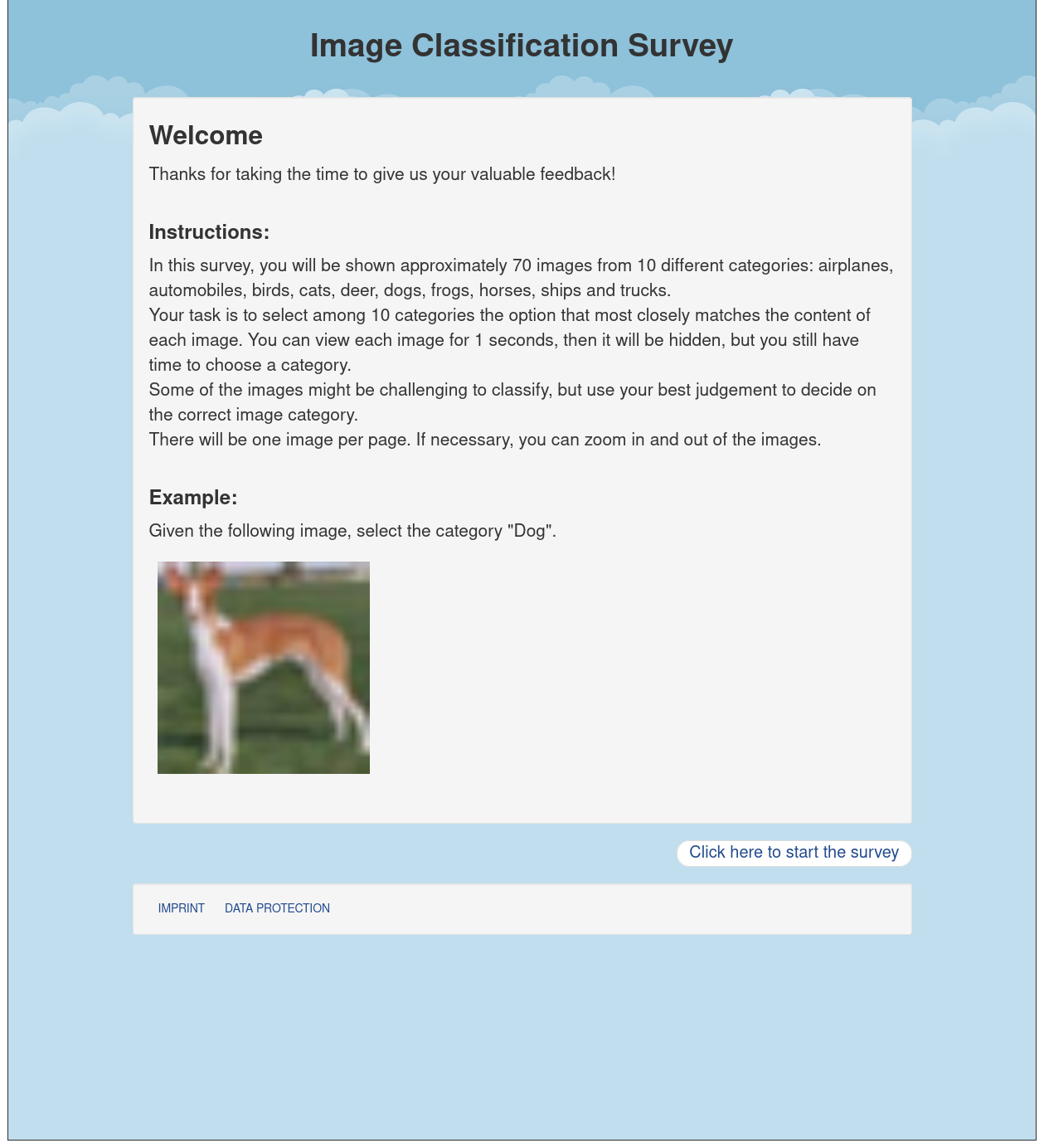}
        \caption{Survey intro}
        \label{fig:error_exp_survey_screenshot_intro}
    \end{subfigure}
    \begin{subfigure}[b]{0.325\textwidth}
        \includegraphics[trim={4cm 1cm 4cm 0cm},clip,width=1\textwidth]{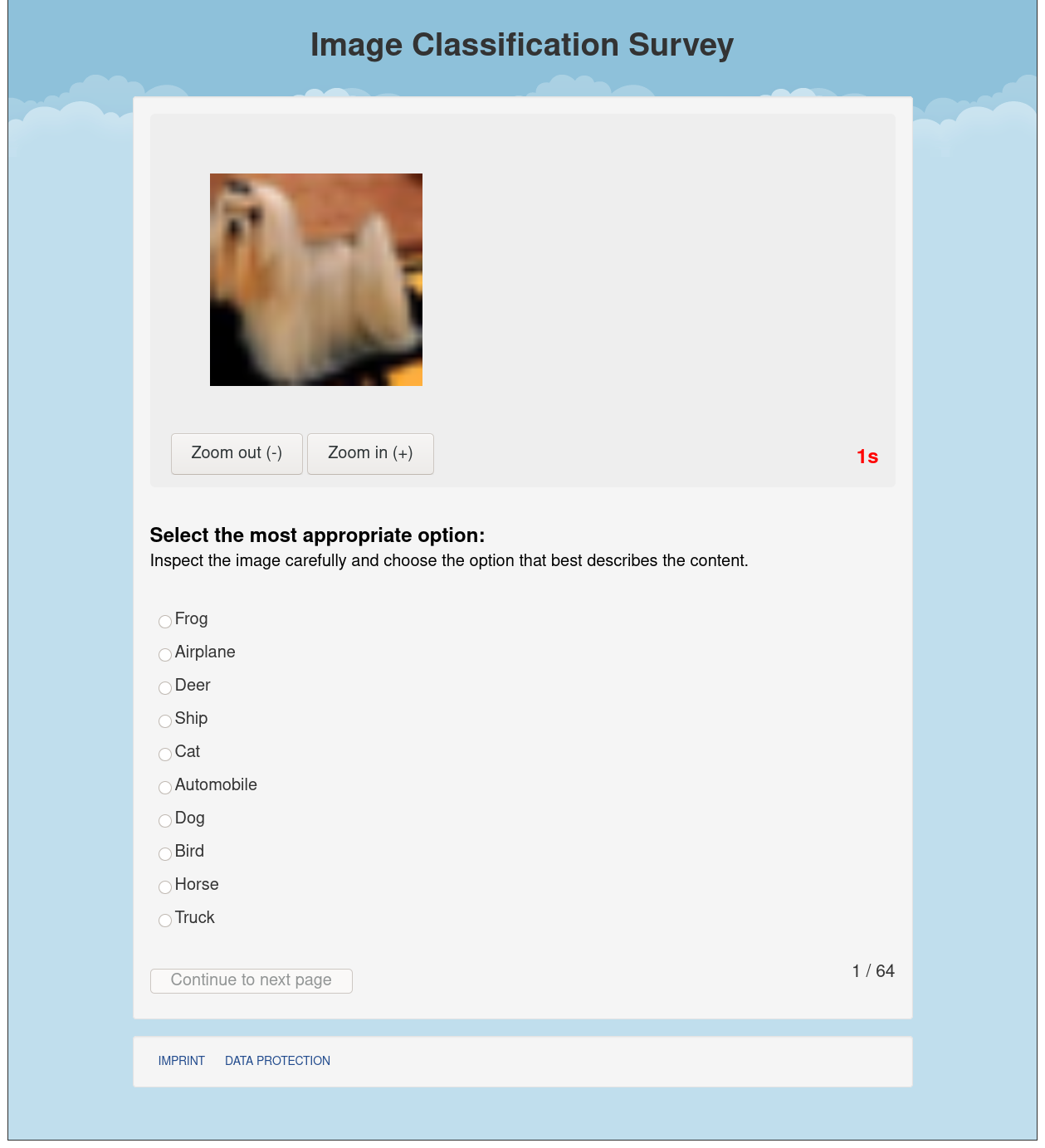}
        \caption{Survey question page}
        \label{fig:error_exp_survey_screenshot_question}
    \end{subfigure}
    \begin{subfigure}[b]{0.325\textwidth}
        \includegraphics[trim={4cm 1cm 4cm 0cm},clip,width=1\textwidth]{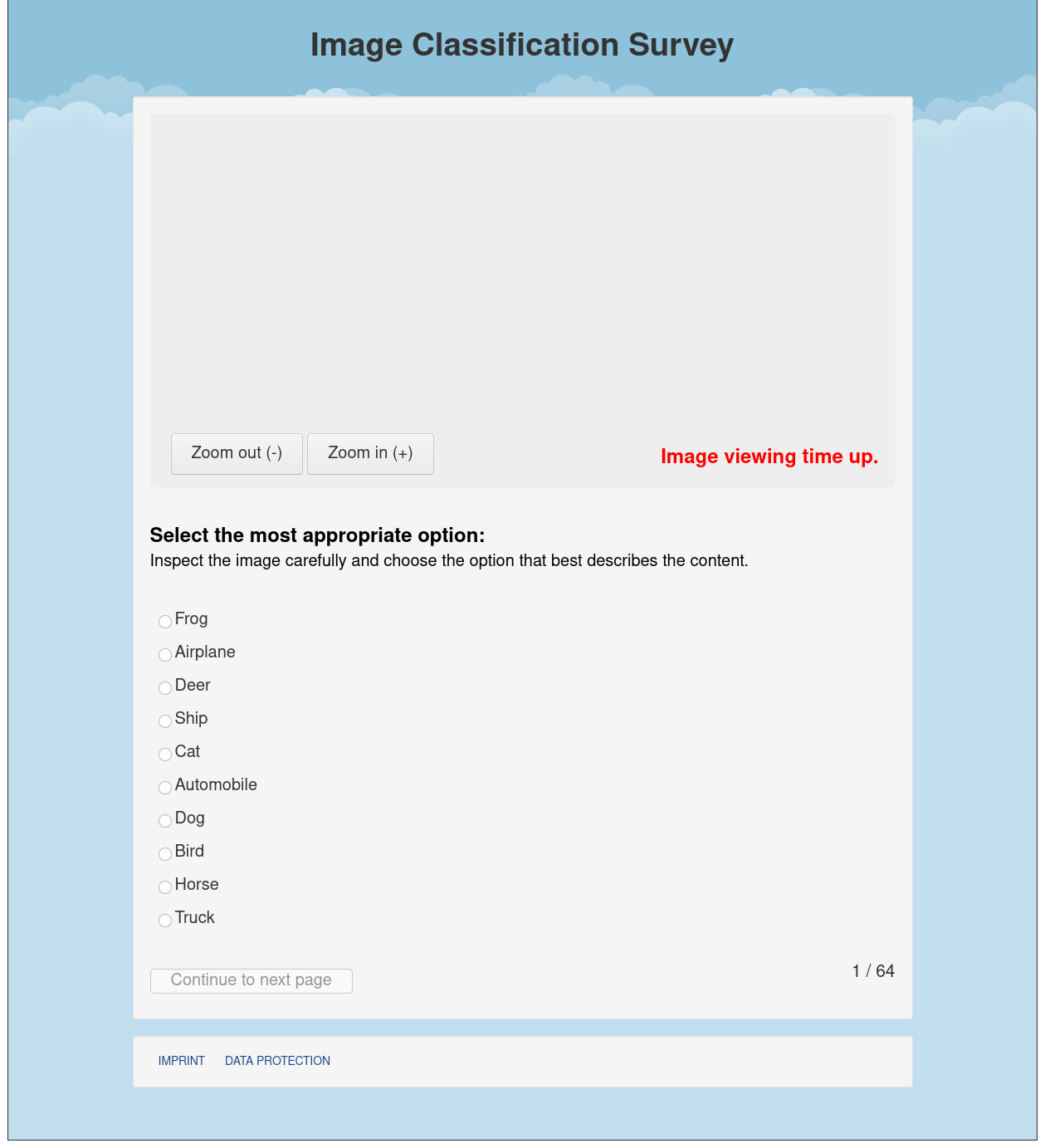}
        \caption{Survey question page -- timed out}
        \label{fig:error_exp_survey_screenshot_question_to}
    \end{subfigure}
\caption{ \textbf{[Error Explainability Survey, Screenshots]} 
    Figure~\ref{fig:error_exp_survey_screenshot_intro} shows the survey's intro page containing instructions for the participants.
    Figure~\ref{fig:error_exp_survey_screenshot_question} shows the web app interface used by participants to view and classify images and Figure~\ref{fig:error_exp_survey_screenshot_question_to} shows how the interface changes after the 1s viewing time has expired.}
\label{fig:error_exp_survey_screenshots}
\end{figure}

\setlength{\tabcolsep}{0.5em}
\begin{table}[ht]
    \centering
    \caption{\textbf{[Human accuracy for different viewing duration timeout values]}
    As the viewing time is reduced from 5 to 1 seconds in our prior studies we see a decrease in human classification accuracy.
    This reduction, however, is small at only about $1.6\%$ from 5 to 1 seconds, so we choose an image viewing timeout of 1 second in our main survey.
    }
    \begin{tabular}{c||c}
        \hline
        \thead{Viewing duration timeout} & \thead{Human Accuracy} \\
        \hline
        \thead{5s} & $93.4\%$ \\
        \thead{3s} & $92.2\%$ \\
        \thead{2s} & $91.8\%$ \\
        \thead{1s} & $91.8\%$ \\
        \bottomrule
    \end{tabular}
    \label{tab:error_exp_timeout_validation_results}
\end{table}

\xhdr{Survey setup.}
We run a survey where the task is to classify the given image into one of the 10 classes in the data.
We recruit 150 workers from Amazon Mechanical Turk (AMT) for the survey.
To keep the workload for each worker manageable, we create multiple random partitions of the set of 344 images into five subsets, four with size 70 and one with size 64, such that each subset contains 10 randomly chosen images from each of the C+, C- and NC-6-12 categories, between 11 and 15 images from NC<6 and between 22 and 25 images from NC>12 (the goal was to show each category to a similar number of workers, and also to ensure that each worker sees a proportional fraction of images from each category).

After an introduction page containing instructions for the image classification task, we show one of the randomly chosen subsets of images to each worker in a web app which we created for the study.

By default, we enlarge each image from its original 32x32 pixel size to 256x256 pixels, however, the workers could zoom out the image to the original size or further zoom into the image.
Screenshots of the web app interface seen by the AMT workers are shown in Figure \ref{fig:error_exp_survey_screenshots}.
As a result of assigning workers to randomly chosen subsets of images, we obtain a varying number of responses per image, ranging from 20 to 40, with an average of 30 responses per image.

\interfootnotelinepenalty=10000

To understand which images are \textit{easy} for humans to classify, we limit the time that each worker can view each of the images.%
\footnote{This choice is based on the well-known phenomenon in psychology and neuroscience called the speed-accuracy tradeoff.
We hypothesize that images which are easy to classify for humans do not require a lot of time to make a decision and therefore setting a limit on viewing time enables us to distinguish between images that are easy and hard to classify for humans.
See: Heitz RP. The speed-accuracy tradeoff: history, physiology, methodology, and behavior. Front Neurosci. 2014;8:150. Published 2014 Jun 11. doi:10.3389/fnins.2014.00150}
However, after an image is hidden, there is unlimited time to choose among the 10 categories.

We assume that correct and explanation-conformant images which make up $84.6\%$ of the dataset are most representative of easy to classify images and therefore use them to calibrate the appropriate amount of viewing time necessary for correctly classifying an image.
To validate the viewing duration value, we conducted a four-part prior study on a randomly chosen set of 50 correct and explanation-conformant images not used in the main study, which we showed for 5, 3, 2 and 1 seconds to 25 AMT workers each and asked them to choose the correct class, while keeping all other parameters the same as in the main study.
We find only small differences in performance for the four viewing durations, as shown in Table \ref{tab:error_exp_timeout_validation_results}.
Consequently, use a viewing duration value of 1 second in our main study.

\xhdr{Workers and compensation.}
For each survey, we recruit 25 workers from AMT.
We only admit workers (i) from the US, who (ii) have the \text{master} qualification, (iii) have at least 95\% previous HIT approval rate, and (iv) at least 100 approved assignments on AMT.
The compensation was set to 8 USD per participant.
The average completion time of the survey was less than 25 minutes.

\subsection{Measures}
\label{sec:error_exp_measures}

\xhdr{Human Accuracy.}
We measure the average human accuracy, \ie~the fraction of workers who choose the correct category for each image, averaged over all images.

\xhdr{Human confusion via Shannon Entropy.}
We use Shannon Entropy as a measure of confusion among the humans in predicting the class of the image.

We randomly subsample the votes for each image to the minimum number of responses of 20 for any image to make the entropy computationally comparable.
We repeat this process 10 times with different random seeds to obtain robust results.
If our entropy-based confusion measure is high for an image it means, that votes for the different classes are distributed relatively uniformly and therefore there is high confusion among humans.

\xhdr{Human Agreement.}
To measure how much humans agree on predicting the class of a given  image, we compute what share of all votes is allocated to the class that receives the majority of all votes.
In the CIFAR-10 data with 10 classes, completely random votes would results in a human agreement of $10\%$, whereas all votes for the same class would results in a human agreement of $100\%$.

\xhdr{Reweighting}
The set of images used in the survey is made up of five types of images, C+, C-, NC<6, NC-6-12 and NC>12, which are sampled out of proportion from the original dataset (in order to ensure ample participation from each category).
Therefore, when we report results for images from multiple of these groups such as in Table \ref{tab:error_exp_agg_performance}, we compute the metrics for each set of images separately and then perform a weighted aggregation according to how prevalent each of the groups is in the original CIFAR-10 test set is.
The prevalence of each group of images can be seen in Table \ref{tab:error_exp_data_fractions}.

\subsection{Human Experiment Results}
\label{sec:appendix_error_explainability_results}

\begin{table}[t]
    \centering
    \caption{\textbf{[Human performance on the different groups of images used in the survey]}
    Humans perform better -- \ie~have higher accuracy and lower confusion -- on images that are correct and explanation conformant, which are also the images that are easy for the machine to classify.
    Additionally, as the number of detected \concepts increases, humans are less confused, even though human accuracy is similar.
    This correlates with an increase in machine accuracy.
    }
    \begin{tabular}{c||c|c|c}
        \hline
        \multicolumn{1}{c}{} & \thead{Human Accuracy} & \thead{Human Confusion \\ (Shannon Entropy)} & \thead{Machine Accuracy} \\
        \hline
        \thead{Correct and \\ Explanation \\ Conformant (C+)} & $91.8\%$ & $0.212$ & $100.0\%$ \\
        \thead{Incorrect and \\ Explanation \\ Conformant (C-)} & $83.7\%$ & $0.407$ & $0.0\%$ \\
        \thead{Less Than 6 \\ \concepts (NC<6)} & $83.1\%$ & $0.418$ & $31.9\%$ \\
        \thead{Between 6 and 12 \\ \concepts (NC-6-12)} & $82.9\%$ & $0.4$ & $42.0\%$ \\
        \thead{Greater Than 12 \\ \concepts (NC>12)} & $84.7\%$ & $0.325$ & $47.5\%$ \\
        \bottomrule
    \end{tabular}

    \label{tab:error_exp_data_split_performance}
\end{table}

\setlength{\tabcolsep}{0.5em}
\begin{table}[t]
    \centering
    \caption{\textbf{[Human performance on different divisions of the dataset]}
    Humans exhibit higher classification accuracy and lower confusion on explanation conformant and on images correctly classified by the machine, compared to explanation non-conformant and incorrectly classified images, respectively.
    However, for explanation conformance this gap is larger for both metrics than for machine correctness.
    The human and machine performance numbers are extrapolated to the entire CIFAR-10 test set by reweighting the performance achieved on each of the image partitions according to their prevalence in the test set (see Table \ref{tab:error_exp_data_fractions}).
    }
    \begin{tabular}{c||c|c|c}
        \hline
        \multicolumn{1}{c}{} & \thead{Reweighed \\ Human Accuracy} & \thead{Reweighed \\ Human Confusion \\ (Shannon Entropy)} & \thead{Reweighed \\ Machine Accuracy} \\
        \hline
        \thead{All Images} & $90.5\%$ & $0.24$ & $88.3\%$ \\
        \hline
        \thead{Explanation \\ Conformant} & $91.3\%$ & $0.215$ & $92.8\%$ \\
        \thead{Explanation \\ Non-Conformant} & $83.2\%$ & $0.387$ & $42.0\%$ \\
        \hline
        \thead{Machine Correct} & $90.7\%$ & $ 0.233$ & $100.0\%$ \\
        \thead{Machine Incorrect} & $85.0\%$ & $0.388$ & $0.0\%$ \\
        \bottomrule
    \end{tabular}
    \label{tab:error_exp_agg_performance}
\end{table}

\begin{figure}[h]
    \centering
    \begin{subfigure}[b]{0.47\textwidth}
        \includegraphics[trim={0cm 0cm 0cm 0cm},clip,width=1\textwidth]{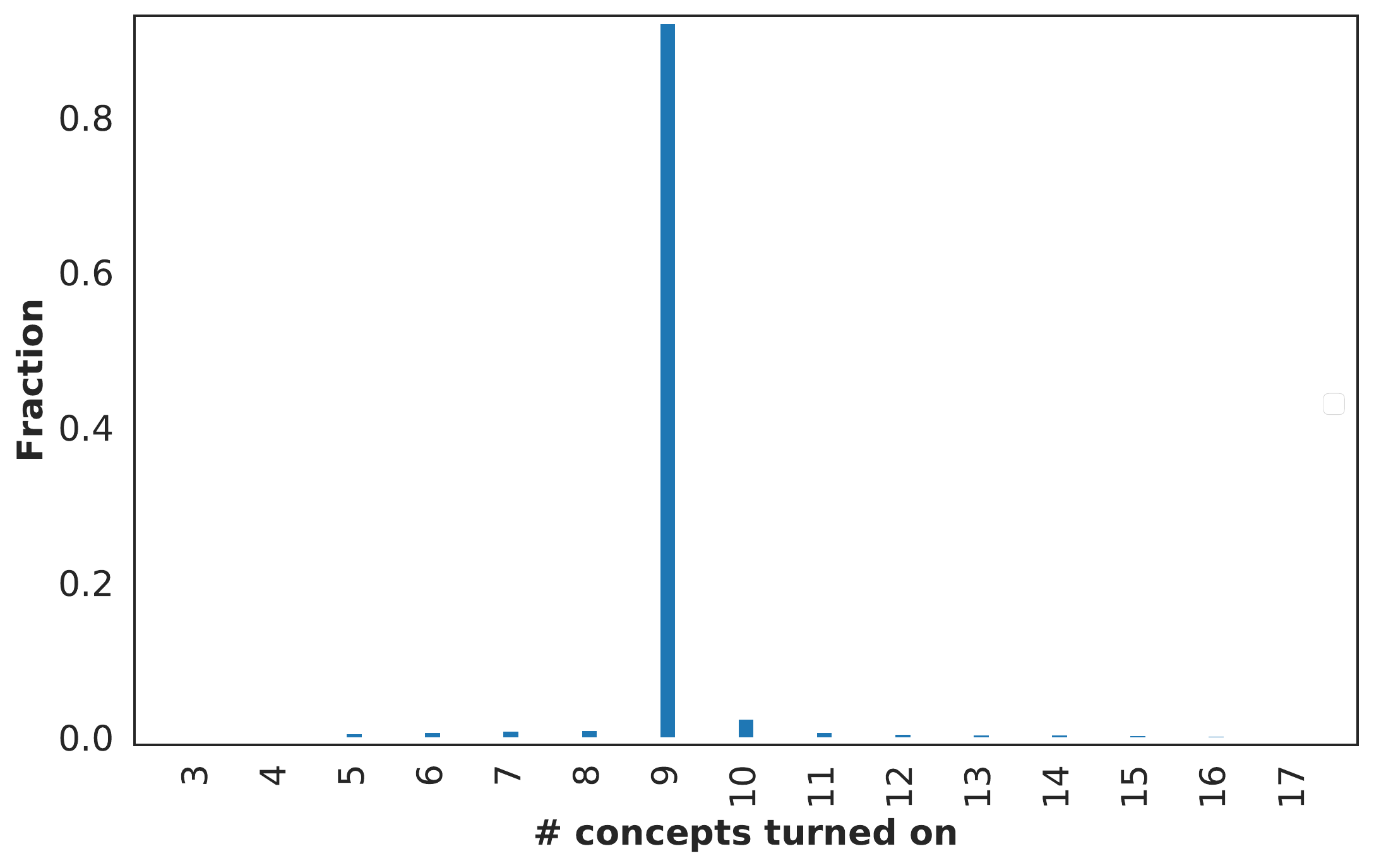}
        \caption{Overall.}
        \label{fig:num_concepts_turned_on_overall}
    \end{subfigure}
    \begin{subfigure}[b]{0.47\textwidth}
        \includegraphics[trim={0cm 0cm 0cm 0cm},clip,width=1\textwidth]{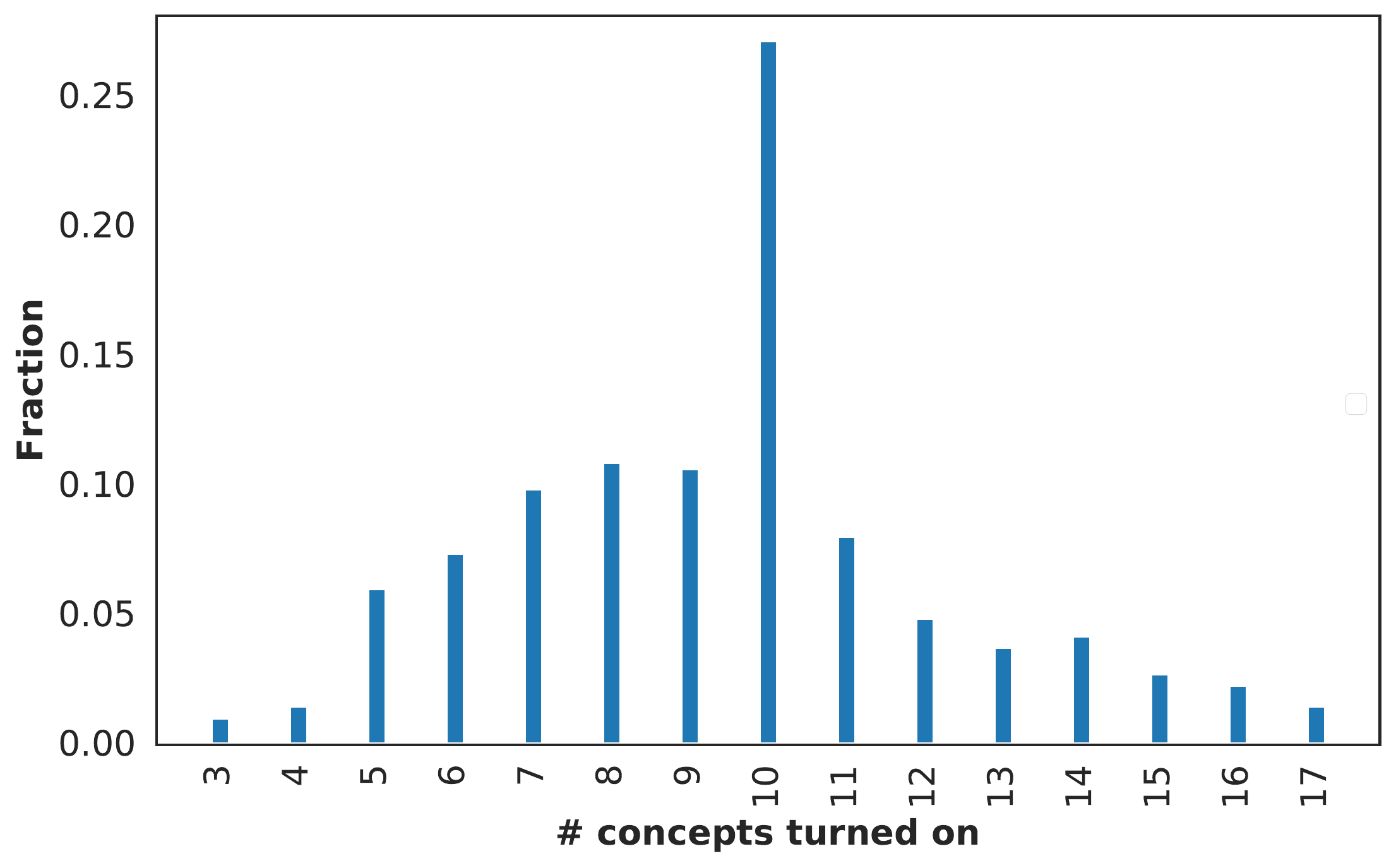}
        \caption{Non explanation-conformant predictions.}
        \label{fig:num_concepts_turned_on_inconsistent}
    \end{subfigure}
\caption{ \textbf{[\# \conceptlabels turned on, CIFAR-10]} Figure~\ref{fig:num_concepts_turned_on_overall} shows the number of \conceptlabels detected for the entire dataset. The peak at 9 \conceptlabels aligns with our observation that most of the predictions (about $91\%$, Table~\ref{table:error_estimation}) are explanation-conformant and hence have exactly 9 \conceptlabels detected. Figure~\ref{fig:num_concepts_turned_on_inconsistent} shows the distribution for just non explanation-conformant predictions. Notice that there are a few samples where the model detected very few (<6) and very large (>12) number of \conceptlabels.
}
\label{fig:num_concepts_turned_on}
\end{figure}

\begin{figure}[t]
\centering
\includegraphics[width=0.6\textwidth]{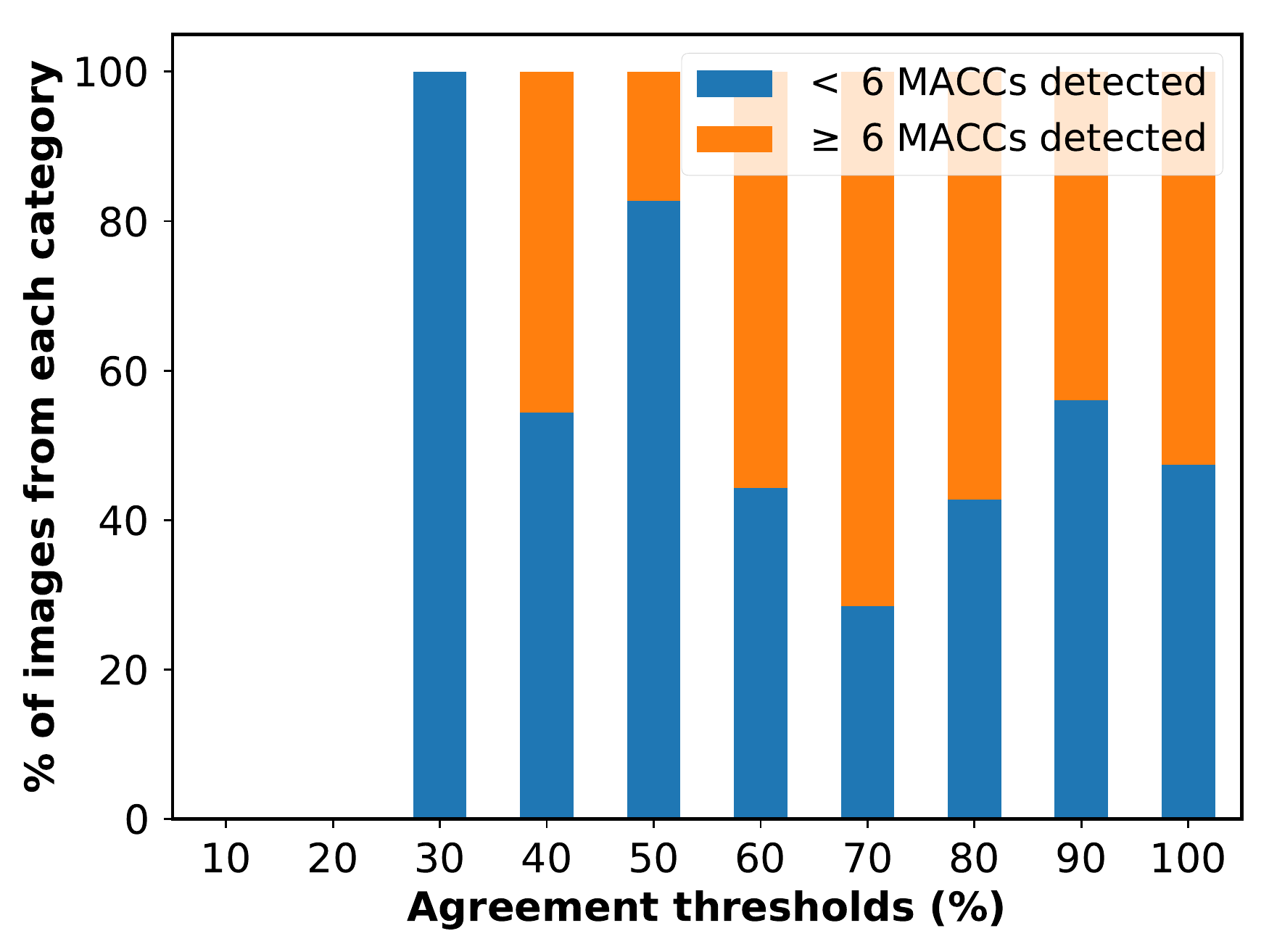}
\caption{ \textbf{[Human agreement in image classification for explanation non-conformant images]}
    Each agreement threshold value corresponds to a bin containing explanation non-conformant images where human agreement is in the range of $(10\%, 20\%]$, $(20\%, 30\%]$, etc. where the values on the x-axis mark the upper boundary of the bins.
    For each bin we show its composition in terms of images with less than 6 and 6 or more detected \concepts by the machine.
    Since there are $172$ images with 6 or more \concepts but only $72$ images with less than 6 \concepts detected, we normalize the contributions of each of the groups such that if a bin would contain less than 6 and 6 or more detected \concept images in a ratio of $72:172$, we would show a $50\%$ share each. \\
    The figure shows, that images with low human agreement mostly are ones with less than 6 detected \concepts.
    This is an indication, that human agreement and perception and \concept detection are correlated.
    }
\label{fig:error_exp_agreements}
\end{figure}

\begin{figure}[t]
\centering
\includegraphics[width=1.0\textwidth]{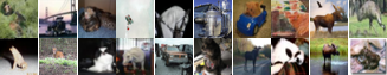}
\caption{ \textbf{[Images with least human agreement]}
    Explanation non-conformant images that exhibit the least agreement among humans.
    The images are ordered with agreement increasing from left to right, top to bottom.
    }
\label{fig:error_exp_low_plurality_images_extended}
\end{figure}

First, we measure human accuracy and confusion on each of the five subsets of images used in the survey and report the results in Table \ref{tab:error_exp_data_split_performance}.
For comparison we report machine accuracy as well.
We observe that humans are considerably more accurate and show less confusion on correct and explanation conformant images, indicating that there seems to be an overlap between the images deemed easy by humans and on which the machine performs well.
Another finding is that for non-conformant images, although exhibiting similar error rates, humans seems to be less confused as the number of detected \concepts increases.
This correlates with increased machine accuracy as well.

In Table \ref{tab:error_exp_agg_performance} we further analyze human classification performance by grouping images based on explanation conformity and also on whether they are correctly classified by the DNN. Additionally we show performance on the entire set of 344 images.
We observe that humans make more accurate classifications and show less confusion on explanation conformant and correctly classified images, compared to non-conformant and incorrectly classified images respectively.
For explanation conformity, however, the accuracy and confusion differences are larger then for correctness, meaning that explanation conformity seems to be a better indicator of difficulty in image classification for humans compared to correctness of machine predictions.

Finally, to further examine the link between the number of detected concepts and human image classification performance, we test whether as the number of detected \concepts increases, human confusion in classification reduces.
Figure~\ref{fig:num_concepts_turned_on} shows that for some images the machine classifier detects very few concepts.
We divide the set of explanation non-conformant images into ones with less than 6 and 6 or more detected \concepts.
For both sets we measure human agreement as described in section \ref{sec:error_exp_measures}.
Figure \ref{fig:error_exp_agreements} shows that images with low human agreement primarily have less than 6 \concepts detected and images with higher agreement mostly have 6 or more \concepts detected.
This indicates that human agreement and \concept detection are indeed connected.
Figure \ref{fig:error_exp_low_plurality_images_extended} shows the images with the lowest agreement among humans.

\section{Additional experiments on Error Vulnerability}\label{sec:appendix_error_vulnerability}

We use the implementation of FGSM, DeepFool, CarliniWagner and PGD provided by the widely used (and publicly available) repository Foolbox (v1.8.0) \cite{rauber2017foolbox}.

\subsection{Performing an explanation-conformant adversarial perturbation and comparison with \citet{guo2017calibration}}\label{sec:appendix_consistent_attack}

To perform an explanation-conformant adversarial attack, we build on PGD by \citet{aleks2017pgd}. Throughout the description of the algorithm, we use the following notation: $\hat{y}$ is the predicted class label, $y$ is the ground truth, $r$ is the vector of ground truth \conceptlabels, $\hat{r}$ is the vector of predicted \conceptlabels, and $\hat{r}_{const}$ are the \conceptlabels which pass the explanation-conformity check (corresponding to the predicted class) \ie, ($\hat{y}, \hat{r}_{const}$) is an explanation-conformant prediction. $\rho$ is the budget given to the attacker \ie, if the perturbed image is $x'$ and the original image is $x$ then $||x - x'||_2 \leq \rho$.

In the paper we use ``post-hoc'' training as described in section~\ref{sec:training_concepts}. This is a 2 step process, where we first train the model to predict class labels, and in the second step we add a head to predict \conceptlabels, keeping all the previous layers fixed from the first training step.
We denote the neural network until the penultimate layer by $f_{rep}$, the final layer for predicting class labels by $f_{class}$ and the final layer for predicting \conceptlabels by $f_{concept}$. Under our framework, for a given input $x$, class label prediction is given by $f_{class}(f_{rep}(x))$ and \concept prediction is given by $f_{concept}(f_{rep}(x))$ (notice that class and \conceptlabels predictions both share common hidden layer representations via $f_{rep}$). The losses for class and \conceptlabels are given by $L_{class}(f_{class}(f_{rep}(x)))$ (we use cross-entropy loss in our implementation) and $L_{concept}(f_{concept}(f_{rep}(x))$ (we use binary cross entropy loss in our implementation). Given these notations, Algorithm \ref{algo:pgd_consistent} proposes a method to generate explanation-conformant adversarial samples. Note that the algorithm is same as that of \citet{aleks2017pgd}, however the only change required to generate explanation-conformant adversarial samples is the change in the loss function.

\begin{algorithm}
 \KwData{Ground-truth class label $y$, predicted class label $\hat{y}$, ground truth \conceptlabels $r$, predicted \conceptlabels $\hat{r}$, input images $x$, $f_{class}$, $f_{concept}$, $f_{rep}$, $\rho$}
 \KwResult{Perturbed image $x'$}
 $x^0 = x$, $k = 0$;\\
 \While{$y == \hat{y}$ || not explanation-conformant$(\hat{r}, \hat{y)}$}{
 $L := L_{class}(f_{class}(f_{rep}(x)), y) - L_{concept}(f_{class}(f_{rep}(x)), \hat{r})$
 \Comment{$\hat{r}$ represents the consistent concept labels and we want to minimize this loss, hence the negative sign}
 $x^{k+1} := x^{k} + \eta * \nabla_x L$;\\
 \If{$|| x^{k+1} = x^0 ||_2 > \rho$}{
    $x^{k+1}$ := projection of $x^{k+1}$ onto the 2-norm ball represented by $|| x^{k+1} - x^0 ||_2 = \rho$;\\
 } 
 $\hat{y} := f_{class}(f_{rep}(x^{k+1})$;\\
 $\hat{r} := f_{concept}(f_{rep}(x))$;\\
 $k := k + 1$;\\
 }
 \caption{Algorithm for generating consistent adversarial samples}
 \label{algo:pgd_consistent}
\end{algorithm}

\setlength{\tabcolsep}{0.5em}
\begin{table*}[ht]
\caption{[Error Vulnerability] Mean L2 distances for attacks on class labels only (\textit{class-only}), on \concept conformity check in addition to the class label (\textit{explanation-conf.}) and on the method of \citet{guo2017calibration} (\citeauthor{guo2017calibration}---see Section~\ref{sec:eval_discussion}). Explanation-conformant attacks require much larger perturbation than the class-only attack and the attack on \citet{guo2017calibration}.}
\centering
{\renewcommand{\arraystretch}{1.25}
\begin{tabular}{lccc}
\hline

 & \multicolumn{1}{c}{\textbf{\begin{tabular}[c]{@{}l@{}}Class-only\end{tabular}}} &  \multicolumn{1}{c}{\textbf{\begin{tabular}[c]{@{}l@{}}Explanation-conf.\end{tabular}}} & \multicolumn{1}{c}{\textbf{\begin{tabular}[c]{@{}l@{}}\citeauthor{guo2017calibration}\end{tabular}}} \\ \hline
\multicolumn{1}{l||}{\textbf{CIFAR-10}} & \multicolumn{1}{c}{0.31 $\pm$ 0.20} & \multicolumn{1}{c}{\textbf{5.31} $\pm$ \textbf{5.62}} & \multicolumn{1}{c}{0.35 $\pm$ 0.20} \\ \hline
\multicolumn{1}{l||}{\textbf{Fashion-MNIST}} & \multicolumn{1}{c}{0.26 $\pm$ 0.14} & \multicolumn{1}{c}{\textbf{3.16} $\pm$ \textbf{2.85}} & \multicolumn{1}{c}{0.49 $\pm$ 0.22}\\ \hline
\end{tabular}
}
\label{table:consistent_attack}
\end{table*}

\begin{figure}[ht]
    \centering
    \begin{subfigure}[b]{0.45\textwidth}
        \includegraphics[trim={0cm 0cm 0cm 0cm},clip,width=1\textwidth]{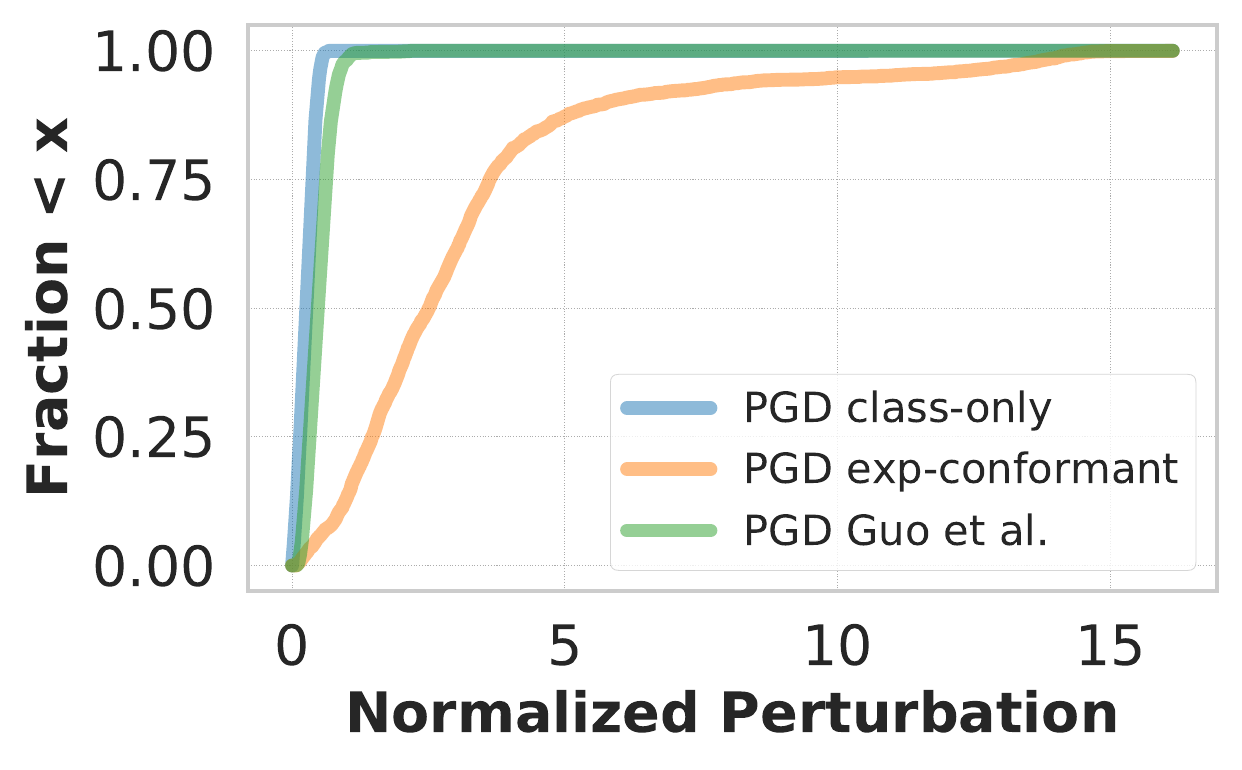}
        \caption{Fashion-MNIST.}
        \label{fig:error_vulnerability_fmnist}
    \end{subfigure}
    \begin{subfigure}[b]{0.45\textwidth}
        \includegraphics[trim={0cm 0cm 0cm 0cm},clip,width=1\textwidth]{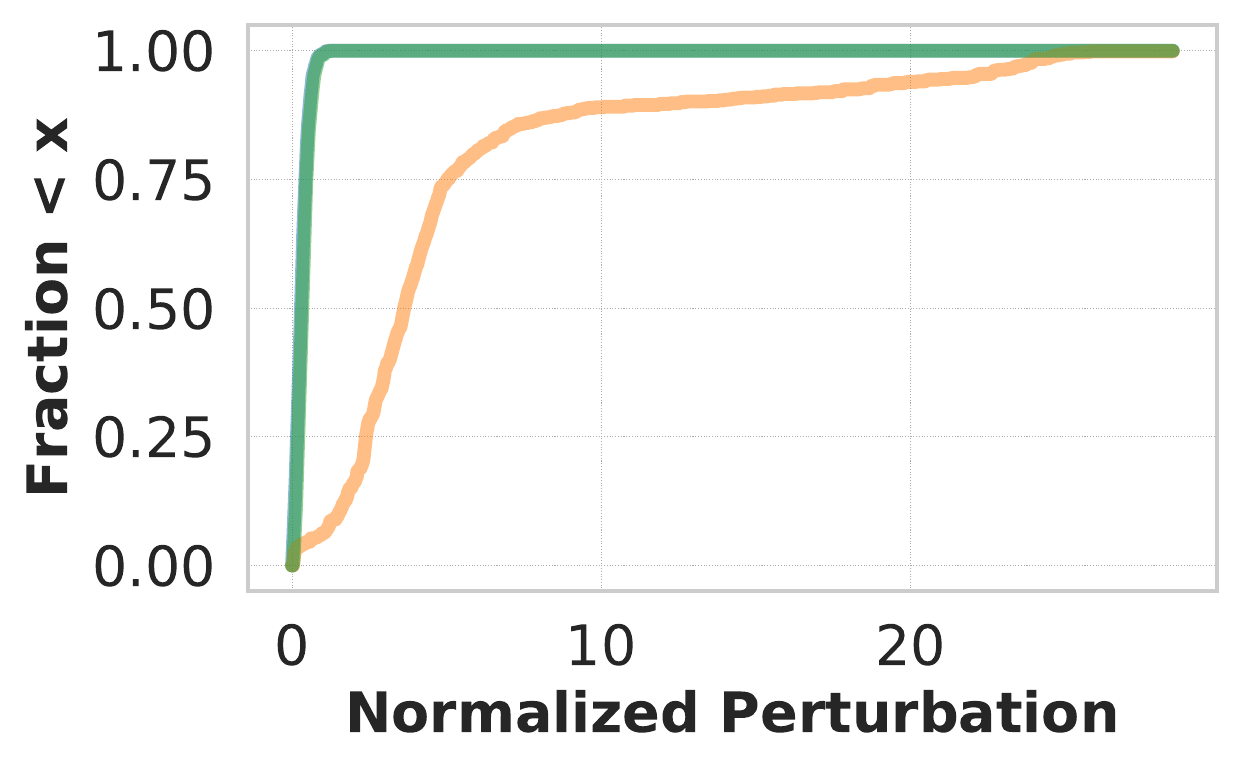}
        \caption{CIFAR-10.}
        \label{fig:error_vulnerability_cifar10}
    \end{subfigure}
\caption{ [Error Vulnerability] Shows the entire distribution of (L2) perturbations in Table~\ref{table:consistent_attack}. In order to attack the explanation-conformity check, the required perturbations are much larger than that of class only and \citeauthor{guo2017calibration}.
}
\label{fig:error_vulnerability}
\end{figure}

\begin{figure}[t]
    \centering
    \begin{subfigure}[b]{\textwidth}
        \includegraphics[trim={3.5cm 8cm 3.5cm 8cm},clip,width=1\textwidth]{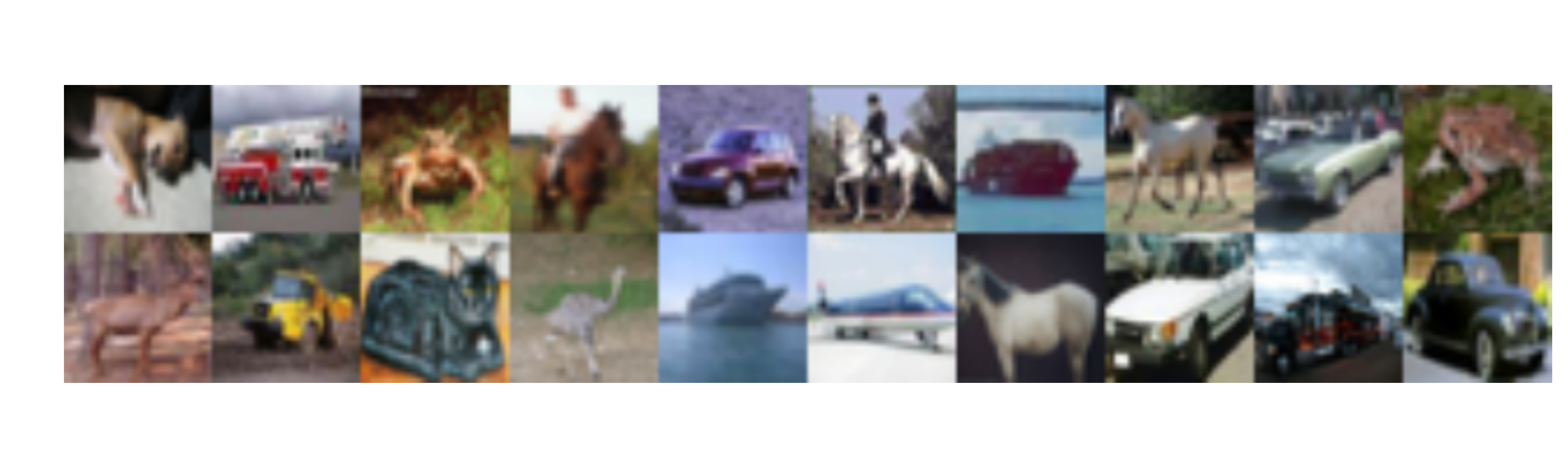}
        \caption{Original Images}
        \label{fig:original_cifar10_all}
    \end{subfigure}
    \begin{subfigure}[b]{\textwidth}
        \includegraphics[trim={3.5cm 8cm 3.5cm 8cm},clip,width=1\textwidth]{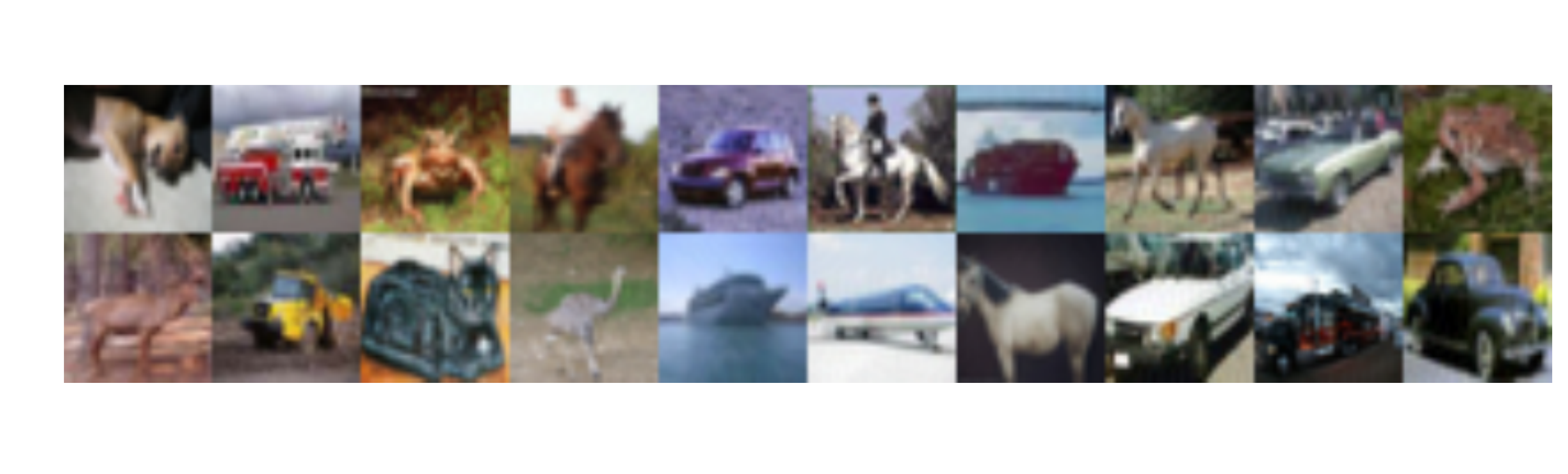}
        \caption{PGD class-only}
        \label{fig:pgd_cifar10_all}
    \end{subfigure}
    \begin{subfigure}[b]{\textwidth}
        \includegraphics[trim={3.5cm 8cm 3.5cm 8cm},clip,width=1\textwidth]{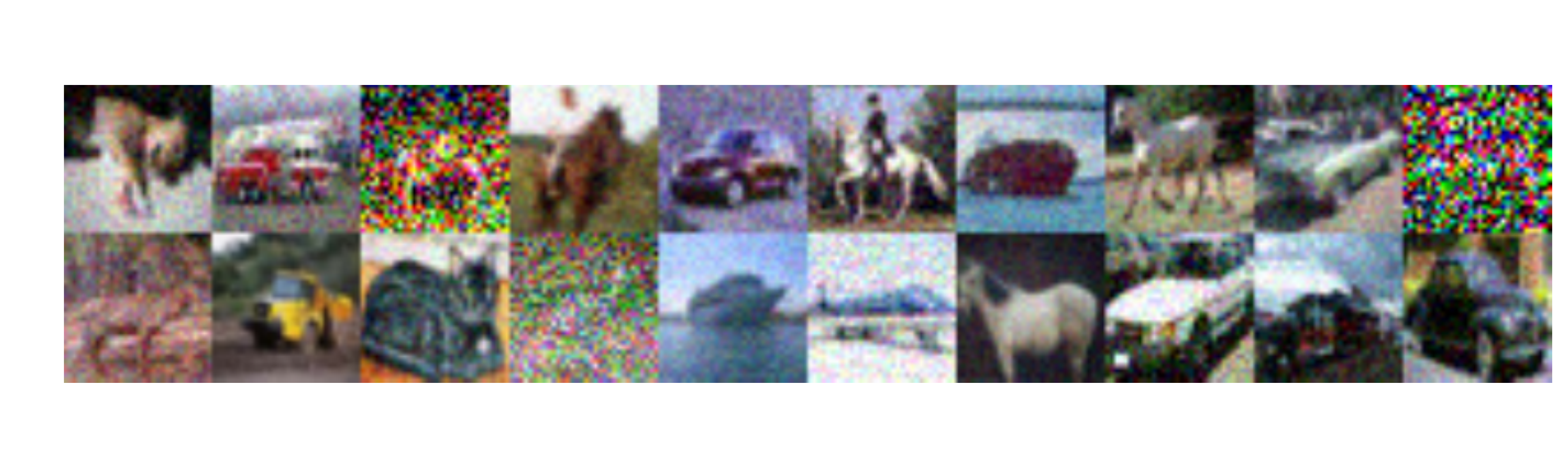}
        \caption{PGD explanation-conformant}
        \label{fig:pgd_consistent_cifar10_all}
    \end{subfigure}
\caption{ [CIFAR-10] Visual quality of samples generated by explanation-conformant PGD attack compared to the traditional PGD attack (changes class label only).
} 
\label{fig:pgd_consistent_images_cifar_all}
\end{figure}

\begin{figure}[t]
    \centering
    \begin{subfigure}[b]{\textwidth}
        \includegraphics[trim={3.5cm 8cm 3.5cm 8cm},clip,width=1\textwidth]{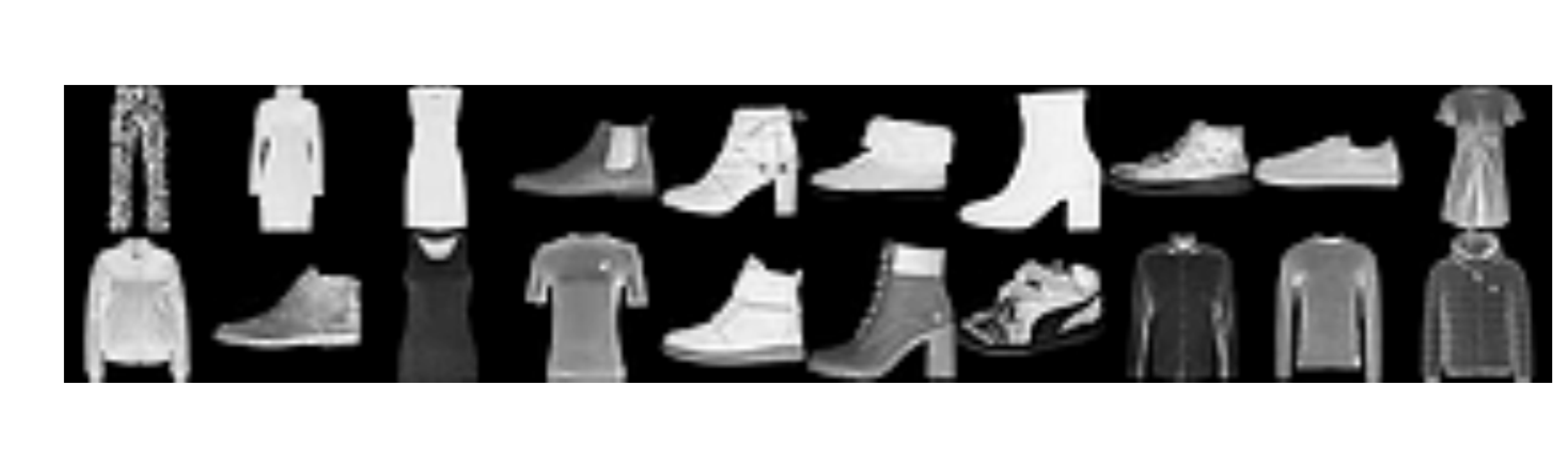}
        \caption{Original Images}
        \label{fig:original_fmnist_all}
    \end{subfigure}
    \begin{subfigure}[b]{\textwidth}
        \includegraphics[trim={3.5cm 8cm 3.5cm 8cm},clip,width=1\textwidth]{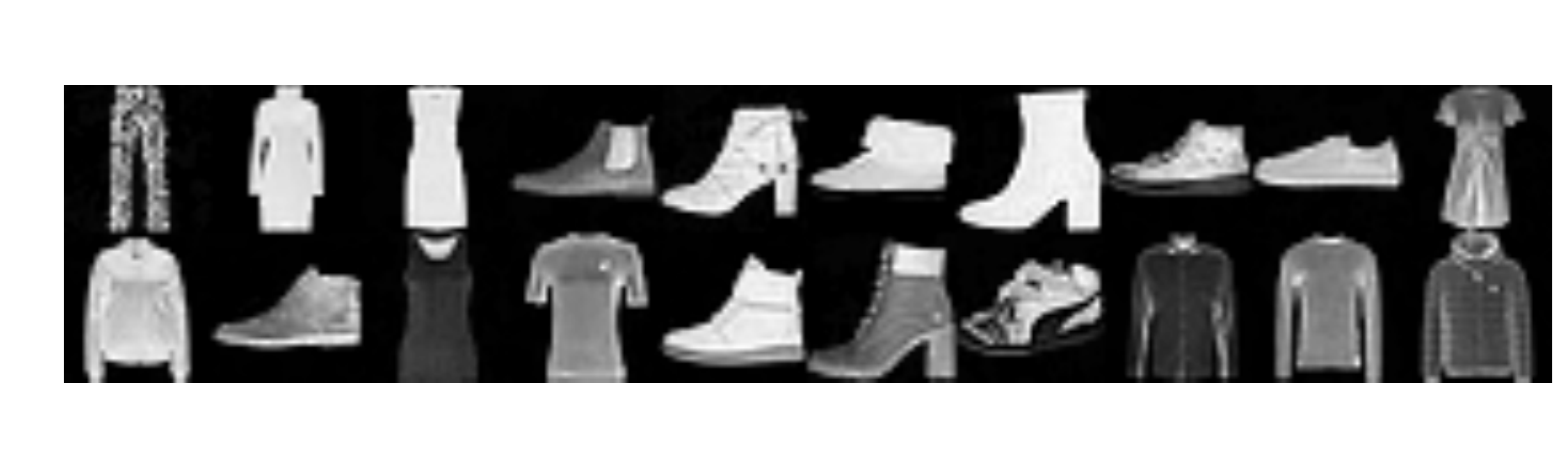}
        \caption{PGD class-only}
        \label{fig:pgd_fmnist_all}
    \end{subfigure}
    \begin{subfigure}[b]{\textwidth}
        \includegraphics[trim={3.5cm 8cm 3.5cm 8cm},clip,width=1\textwidth]{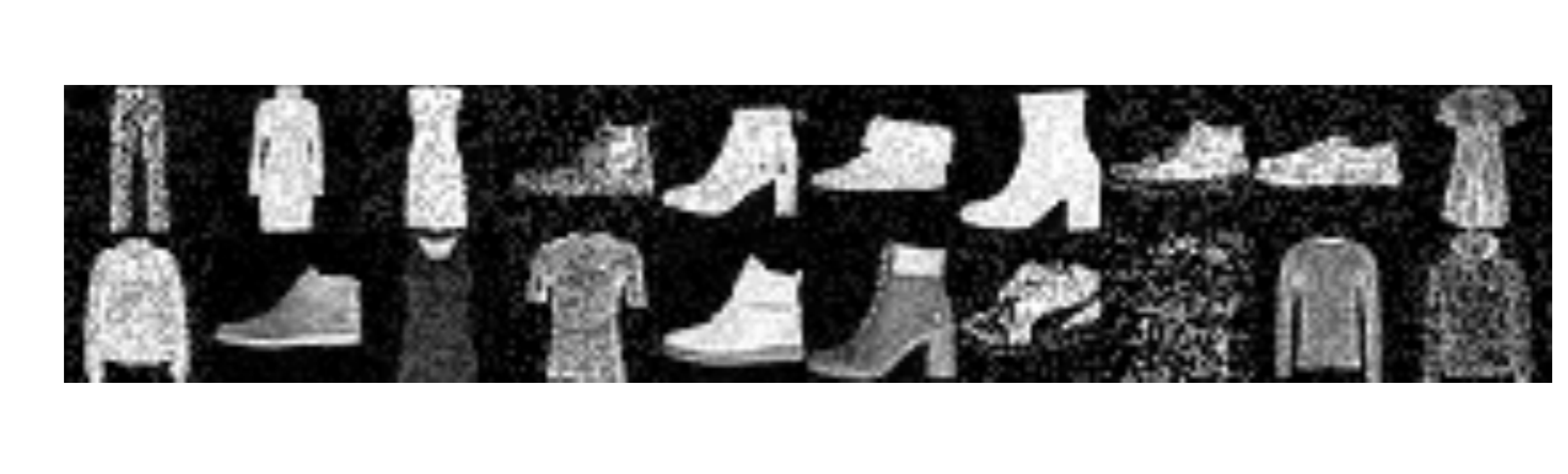}
        \caption{PGD explanation-conformant}
        \label{fig:pgd_consistent_fmnist_all}
    \end{subfigure}
\caption{ [Fashion-MNIST] Visual quality of samples generated by explanation-conformant PGD attack compared to the traditional PGD attack (changes class label only).
} 
\label{fig:pgd_consistent_images_fmnist_all}
\end{figure}

\xhdr{Results.}
As discussed in Section~\ref{sec:error_vulnerability},
Figure~\ref{fig:error_vulnerability} and Table~\ref{table:consistent_attack} show that the amount of perturbation required for explanation conformant attacks is much higher than the traditional attack on just class labels.
Moreover, attacking the calibrated probability-based method of \citet{guo2017calibration} also requires order of magnitude lesser perturbation (than \concept explanation-conformant attack), thus showing that adding \conceptlabels lead to higher robustness against adversarial attacks. Figures~\ref{fig:pgd_consistent_images_cifar_all} \&~\ref{fig:pgd_consistent_images_fmnist_all} show some explanation-conformant adversarial samples generated using Algorithm~\ref{algo:pgd_consistent} as compared to the traditional class-only PGD attack. The explanation-conformant adversarial samples seem to show a much more visible perturbation. We explore the human perceptability of these perturbations in Appendix~\ref{sec:error_vulnerability_human_survey}.

\subsection{Human perceptibility of explanation-conformant adversarial perturbations}\label{sec:error_vulnerability_human_survey}

Here we describe the human survey conducted in Section~\ref{sec:error_vulnerability}. The goal of the survey was to test whether humans find any difference between the \itbf{class-only} and \itbf{explanation-conformant} perturbations conducted in Section~\ref{sec:error_vulnerability}.
We use Amazon Mechanical Turk (AMT) to recruit participants for the surveys.

\xhdr{Survey data.}
We created two groups of perturbed images (i) class-only and (ii) explanation-conformant. Each group consists of 50 randomly selected images.

\begin{figure}[t]
    \centering
    \begin{subfigure}[b]{0.325\textwidth}
        \includegraphics[trim={5cm 2cm 5cm 0cm},clip,width=1\textwidth]{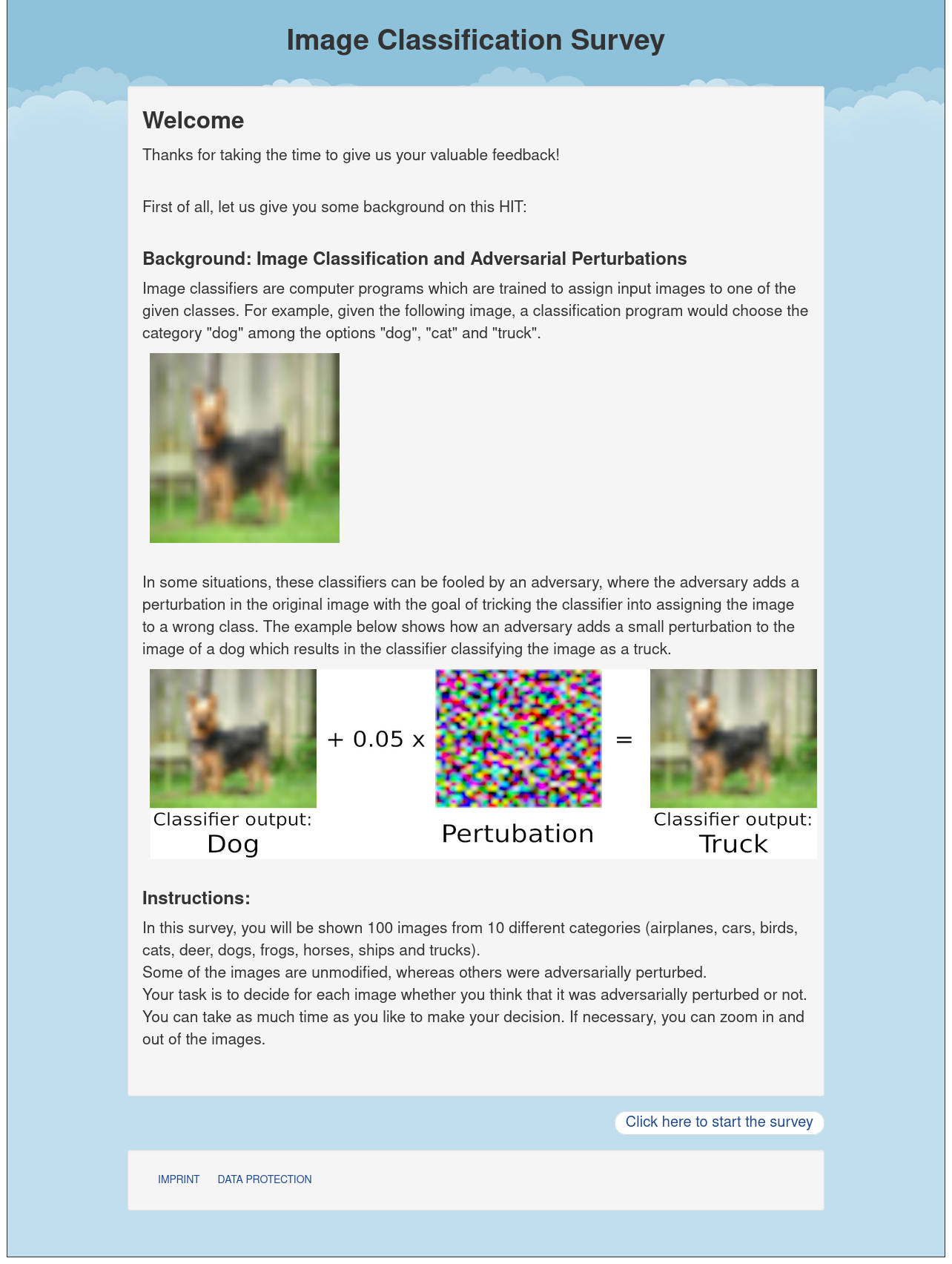}
        \caption{Survey intro \\ \hfill}
        \label{fig:adv_survey_screenshot_individual_intro}
    \end{subfigure}
    \begin{subfigure}[b]{0.325\textwidth}
        \includegraphics[trim={5cm 2cm 5cm 0cm},clip,width=1\textwidth]{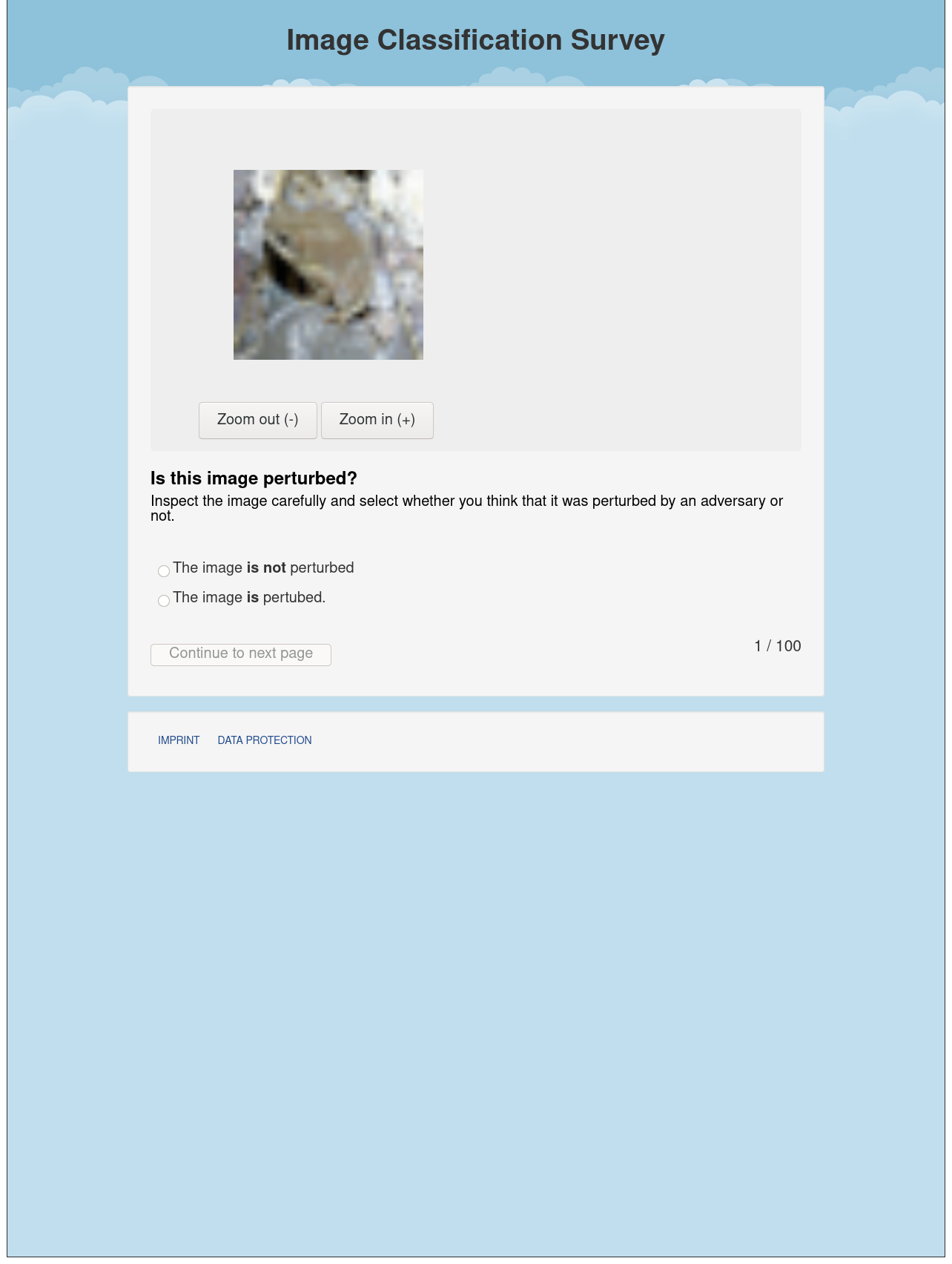}
        \caption{Survey question page: Individual image survey}
        \label{fig:adv_survey_screenshot_individual_question}
    \end{subfigure}
    \begin{subfigure}[b]{0.325\textwidth}
        \includegraphics[trim={5cm 2cm 5cm 0cm},clip,width=1\textwidth]{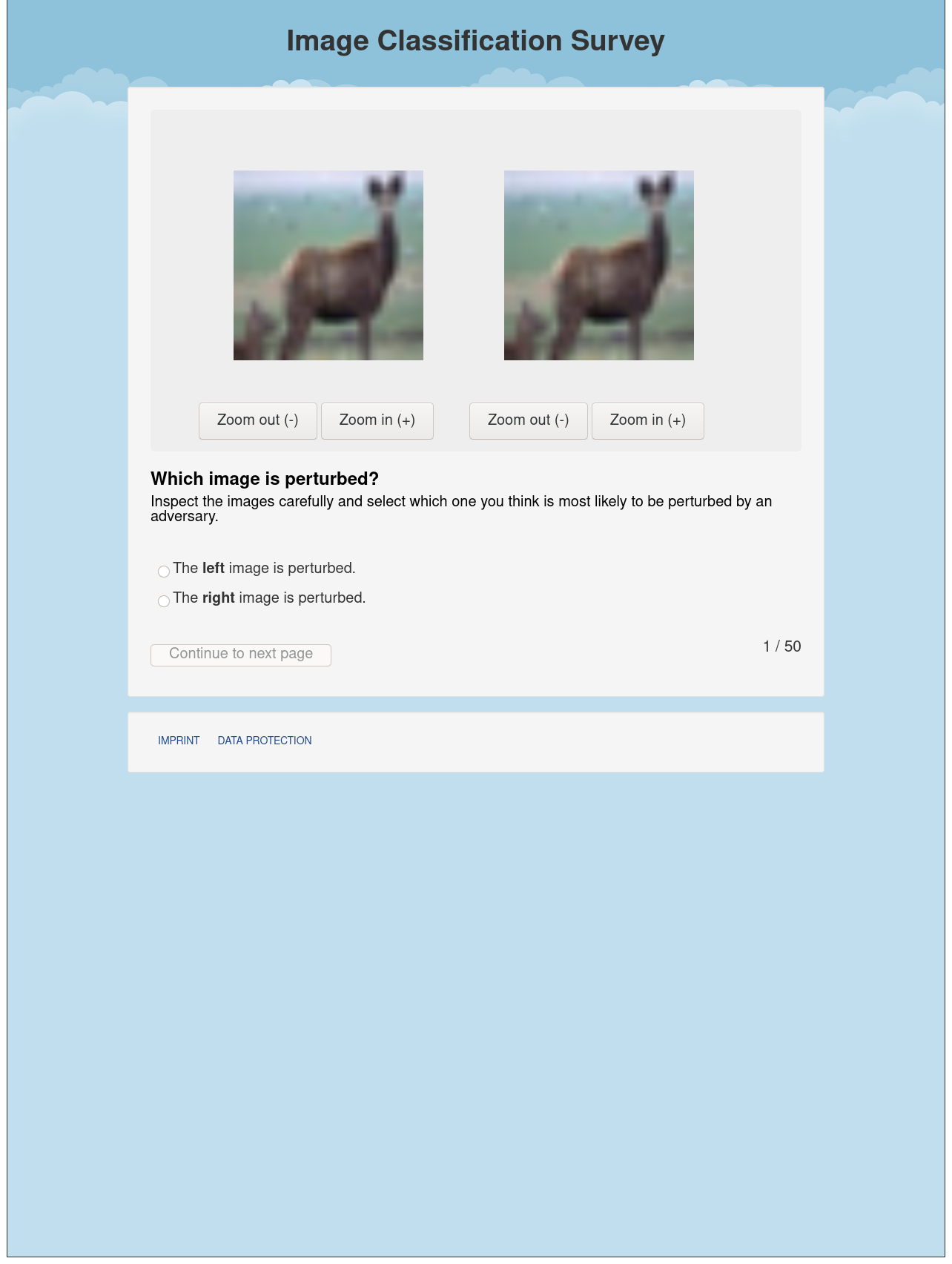}
        \caption{Survey question page: Image-pair survey}
        \label{fig:adv_survey_screenshot_paired_question}
    \end{subfigure}
\caption{\textbf{[Adversarial Survey, Screenshots]}
    Figure~\ref{fig:adv_survey_screenshot_individual_intro} shows the survey's landing page that introduces the workers to the task.
    Figures~\ref{fig:adv_survey_screenshot_individual_question} and~\ref{fig:adv_survey_screenshot_paired_question} show the question pages for the \textit{individual image survey} and \textit{image-pair survey}, respectively. 
    }
\label{fig:adv_survey_screenshots}
\end{figure}

\xhdr{Survey Setup.}
For each of the two sets of attacked images we run two surveys.

\begin{enumerate}
    \item In first \itbf{individual image survey}, we show one image at a time, that is either adversarially perturbed or an unmodified original. The survey consists of 100 images in total: 50 perturbed images and the other 50 being their original counterpart.
    \item In the second survey, we show \itbf{image-pair survey}.
    One image is unmodified and the other one is its adversarially attacked counterpart.
    We then ask each participant for each of the 50 image pairs which of the two images they think was perturbed.
\end{enumerate}

For both surveys, workers had unlimited time to make a decision.
By default, we enlarge each image from its original 32x32 pixel size to 256x256 pixels, however, the workers could zoom out the image to the original size or further zoom into the image.
Screenshots of the web app interfaced shown to participants are shown in Figure \ref{fig:adv_survey_screenshots}.

\xhdr{Workers and compensation.}
For each survey, we recruit 25 workers from AMT.
We only admit workers (i) from the US, who (ii) have the \text{master} qualification, (iii) have at least 95\% previous HIT approval rate, and (iv) at least 100 approved assignments on AMT.
The compensation was set to 8 USD per participant.

\setlength{\tabcolsep}{0.5em}
\begin{table}[t]
    \centering
    \caption{\textbf{[Human performance in adversarial perturbation detection surveys.]}
    The table shows human accuracy in detecting whether an image has been adversarially perturbed or not.
    We report results for two different perturbation types (class-only and explanation-conformant) and two survey types (individual image survey and image pair survey) as defined in Appendix~\ref{sec:error_vulnerability_human_survey}.
    For both survey types, when annotating class-only perturbations, the human detection accuracy is close to random.
    On the other hand, for explanation-conformant perturbations, the human accuracy is much higher.
    Furthermore, humans perform even better on image-pair survey (for both image types).
    }
    \begin{tabular}{c||c}
        \hline
        \thead{Image / Survey Type} & \thead{Human Accuracy} \\
        \hline
        \thead{Class-only / Individual image survey} & $49.8\%$ \\ %
        \thead{Class-only / Image-pair survey} & $58.8\%$ \\
        \hline
        \thead{Explanation-conf. / Individual image survey} & $85.0\%$ \\ %
        \thead{Explanation-conf. / Image-pair survey} & $95.7\%$ \\
        \bottomrule
    \end{tabular}
    \label{tab:adv_survey_performance}
\end{table}

\xhdr{Results.}
The average completion time of the surveys was around 32 minutes.

Table \ref{tab:adv_survey_performance} shows the human accuracy for both survey types and image types. The results show that while the human detection accuracy is close to random for class-only perturbations, it is much higher for explanation-conformant images. This result shows that the explanation-conformant perturbations are indeed so large that humans can detect them in most cases.
Second, we see that images that were attacked in a consistent manner are significantly easier for humans to detect compared to traditionally attacked ones where detection rates are close to random.

\end{document}